\documentclass[11pt]{article}

\usepackage[final]{acl}
\usepackage{times}
\usepackage{latexsym}

\usepackage[T1]{fontenc}

\usepackage[utf8]{inputenc}

\usepackage{microtype}

\usepackage{inconsolata}

\usepackage{graphicx}

\usepackage{algorithm}
\usepackage{enumitem}
\usepackage{newfloat}
\usepackage{listings}
\usepackage[most]{tcolorbox}
\usepackage{xcolor}  
\usepackage{booktabs}
\usepackage{multirow}
\usepackage{amsmath, amssymb}
\usepackage{booktabs}
\usepackage[table]{xcolor}
\usepackage{tabularx}
\usepackage{pifont}
\usepackage{algorithm}
\usepackage{algpseudocode} 

\title{DFAMS: Dynamic-flow guided Federated Alignment based Multi-prototype Search}

\author{
\textbf{
Zhibang Yang\textsuperscript{1,2}\thanks{Equal contribution.},
Xinke Jiang\textsuperscript{1,2}\footnotemark[1],
Rihong Qiu\textsuperscript{1,2,3}\footnotemark[1],
Ruiqing Li\textsuperscript{1,2},
Yihang Zhang\textsuperscript{1},
Yue Fang\textsuperscript{1,2},}\\
\textbf{
Yongxin Xu\textsuperscript{1,3},
Hongxin Ding\textsuperscript{1,2},
Xu Chu\textsuperscript{1,3},
Junfeng Zhao\textsuperscript{1,3}\thanks{Corresponding authors.},
Yasha Wang\textsuperscript{1,2,4}\textsuperscript{\dag}
}
\\[5pt]
\textsuperscript{1}National Engineering Research Center for Software Engineering, Peking University, Beijing, China\\
\textsuperscript{2}School of Computer Science, Peking University, Beijing, China\\
\textsuperscript{3}Key Laboratory of High Confidence Software Technologies, Ministry of Education, Beijing, China\\
\textsuperscript{4}Peking University Information Technology Institute (Tianjin Binhai), Tianjin, China\\
\texttt{\{yangzb, XinkeJiang, RihongQiu, YueFang\}@pku.stu.edu.cn}
}

\begin{document}
\maketitle
\begin{abstract}
Federated Retrieval (FR) routes queries across multiple external knowledge sources, to mitigate hallucinations of LLMs, when necessary external knowledge is distributed. However, existing methods struggle to retrieve high-quality and relevant documents for ambiguous queries, especially in cross-domain scenarios, which significantly limits their effectiveness in supporting downstream generation tasks. Inspired by Dynamic Information Flow (DIF), we propose DFAMS, a novel framework that leverages DIF to identify latent query intents and construct semantically aligned knowledge partitions for accurate retrieval across heterogeneous sources. Specifically, DFAMS probes the DIF in LLMs by leveraging gradient signals from a few annotated queries and employing Shapley value-based attribution to trace neuron activation paths associated with intent recognition and subdomain boundary detection. Then, DFAMS leverages DIF to train an alignment module via multi-prototype contrastive learning, enabling fine-grained intra-source modeling and inter-source semantic alignment across knowledge bases.
Experimental results across five benchmarks show that DFAMS outperforms advanced FR methods by up to 14.37\% in knowledge classification accuracy, 5.38\% in retrieval recall, and 6.45\% in downstream QA accuracy, demonstrating its effectiveness in complex FR scenarios. 
Our code are anonymous available at {\url{https://anonymous.4open.science/r/DFAMS/}}
\end{abstract}

\section{Introduction}
Retrieval-Augmented Generation (RAG) leverages external knowledge documents~\cite{GraphRAG,asai2023retrieval} to effectively enhance the factuality and verifiability of outputs from Large Language Models (LLMs)~\cite{ChatGPT,OpenAI2023GPT4TR,vu2024gptvoicetasker}, significantly mitigating issues such as hallucination and knowledge obsolescence~\cite{10.1145/3571730,cao-etal-2020-factual,HyKGE, jiang2024tc, asai2023selfrag, su2024dragin, jeong2024adaptive, baek2025probing}.
However, mainstream RAG approaches typically rely on a single, centralized vector database for knowledge storage and retrieval
~\cite{kukreja2024performance, bhavnani2009information}. In reality, knowledge is inherently distributed across multiple heterogeneous data sources — for instance, in medical scenarios, retrieval may need to simultaneously access electronic health records (EHR)~\cite{yuan2023ehrdiff}, textbooks, and the latest research papers~\cite{zhao2025medrag}.
\begin{figure}[t]
  \centering
\includegraphics[width=0.4\textwidth]{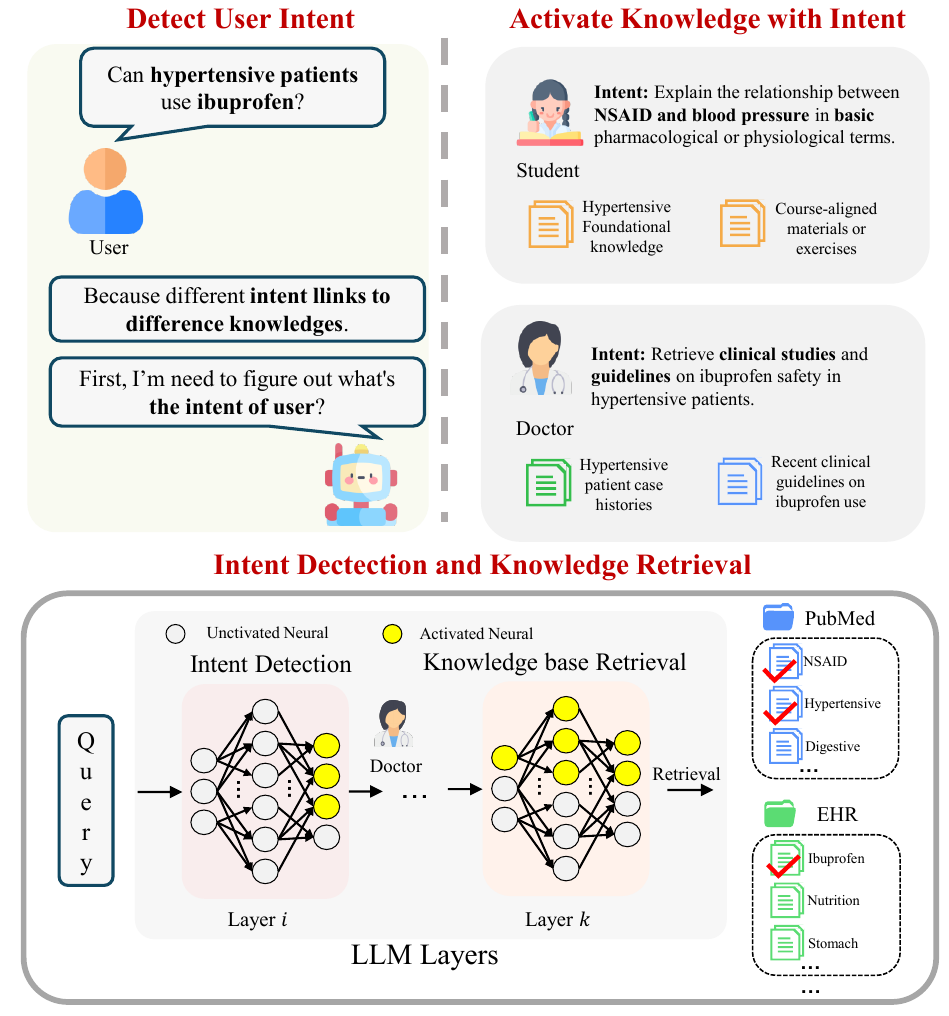}
  \captionsetup{font=footnotesize}
  \caption{Hypothesized process of dynamic information flow (DIF) within LLMs for knowledge base selection. When a user asks “Can hypertensive patients use ibuprofen?”, the LLM first infers the latent intent—where a student seeks basic pharmacological understanding, while a doctor requires clinical evidence. The identified intent (e.g., as a doctor) triggers distinct neural and knowledge activations, forming DIF signals that guide retrieval: clinical pathways access ibuprofen records in EHR, whereas conceptual pathways retrieve NSAID-related information from PubMed.}
   \label{DIF}
   \vspace{-0.6cm}
\end{figure}

Forcibly aggregating all documents into a unified index not only incurs high retrieval costs but also raises concerns around data sovereignty~\cite{jiang2024clinical,shokouhi2011federated, kairouz2021advances}.
To address this, \textbf{Federated Retrieval (FR)} has emerged as a solution, enabling efficient cross-knowledge-base routing decisions that directly and precisely direct queries to the most relevant sources of knowledge~\cite{Scholkopf2019CausalityLearning, guerraoui2025efficient, wang2024feb4rag,ryan2025enronqa, shojaee2025federated}.

Most existing FR methods are primarily designed with an emphasis on routing efficiency, privacy preservation, and downstream task integration~\cite{chakraborty2025federated}.
However, in real-world scenarios with complex semantics~\cite{clarke2008novelty}, users are more concerned with whether the system can accurately retrieve highly relevant documents to effectively support downstream generation tasks~\cite{shokouhi2011federated, huang2024survey}, which does not receive sufficient attention in existing research. 
And due to limitations in current modeling strategies~\cite{wang2024feb4rag}, existing approaches exhibit notable shortcomings in addressing this issue\cite{guerraoui2025efficient}.
On one hand, user queries often suffer from semantic ambiguity or compression, leading to a gap between surface expressions and underlying intent~\cite{yuan2025query}. In different contexts, the same question may require different knowledge sources to answer (as illustrate in Figure \ref{DIF}). 
In such scenarios, user queries often fail to align with the structured and detailed content in the knowledge base, limiting the accuracy and coverage of semantic matching in traditional FR methods~\cite{huang2021efficient}. 
Although some approaches leverage LLMs for prompt-based query rewriting~\cite{hyde} to reduce ambiguity, they often struggle to capture fine-grained semantics due to limited prompt expressiveness
, leading to suboptimal performance in semantically complex scenarios.
On the other hand, in real-world applications 
knowledge bases are typically partitioned by data sources, forming multiple structured yet interrelated knowledge subsystems \cite{wu2025talk}. The semantic boundaries between these subsystems are often flooded and overlapping. 
However, some existing methods 
overlooking underlying semantic connections, 
which struggle to support cross-source recall and dynamic integration.~\cite{wang2024feb4rag}.

Recent research has shown that when LLMs process tasks of different fields, the contribution of each parameter in the LLM model varies~\cite{dhamdhere2018important,yu2018nisp,lecun1989optimal}.
Building on this line of work, further investigations into the structural and functional mechanisms of LLMs have revealed that, during inference, LLMs naturally form a Dynamic Information Flow (DIF)—a latent path through which information propagates dynamically across transformer layers, activating neural substructures associated with semantics, knowledge, and reasoning~\cite{zheng2409attention, yu2024interpreting, wang2024unveiling}. These findings suggest that LLMs may already possess an implicit capability to recognize user intent and organize knowledge into latent subdomains when processing complex queries.
Inspired by this, we raise a central research question: Can DIF within LLMs be explicitly modeled to (1) more accurately identify users’ latent query intents, and (2) structurally segment and dynamically organize overlapping or fuzzy-boundary knowledge subdomains to mitigate semantic misalignment and cross-source retrieval failures in FR settings?
As shown in Figure~\ref{DIF}, we hypothesize that, when presented with a complex query, the LLM first identifies the user's intent. If the model determines that the user is likely a doctor with an intent to retrieve clinical studies and guidelines, it will activate the corresponding neurons, forming a reasoning pathway that generates perception signals related to the target knowledge. These signals may be used to retrieve and align content from heterogeneous sources such as PubMed and EHR, integrating cross-subdomain knowledge for downstream generation.
To validate the above hypothesis, we need to address two core challenges:
(C1) how to accurately detect the relevant DIF within LLMs;
(C2) how to leverage the DIF-based internal semantic organization to support fine-grained knowledge base modeling while preserving semantic associations across multiple sources.
To tackle these challenges, we propose a novel framework named DFAMS (\textbf{D}ynamic-flow guided \textbf{F}ederated \textbf{A}lignment based \textbf{M}ulti-prototype \textbf{S}earch). DFAMS explicitly models the internal DIF of LLMs and constructs knowledge base partitions aligned with the model’s activation patterns, thereby preserving rich semantic signals.
For \textbf{Challenge C1}, we utilize gradient signals under a small number of annotated DIF-probing samples, and apply Shapley value-based attribution methods to identify neuron flow paths associated with query intent recognition and subdomain boundary detection. For \textbf{Challenge C2}, during training, we extract DIF flows induced by queries over each knowledge base. These flows are then used to train an alignment module via multi-prototype contrastive learning, achieving fine-grained intra-source modeling and inter-source alignment. The goal is to maintain semantic continuity across sources while enabling effective knowledge base classification.
In summary, our contributions are as follows:

\begin{itemize}[leftmargin=*,noitemsep,topsep=2pt]
\item 
We reveal a high-dimensional, information-rich DIF in LLMs that encodes both user intent and subdomain knowledge, enabling more faithful query understanding. To our knowledge, this is the first work to exploit DIF for intent-aware, domain-sensitive retrieval modeling.
\item We propose the DFAMS framework. By modeling 
DIF, we construct knowledge partitions that preserve inter-source semantic associations. DFAMS integrates multi-prototype contrastive learning during training and employs 
Adaptive Prototype-Guided Routing 
at inference time, significantly improving the performance.
\item We develop an enhanced FR benchmark 
to encompass realistic and diverse query types, ranging from knowledge-free queries, multi-fragment queries within a single source, to cross-source retrieval. The benchmark integrates structured taxonomy, associated documents, user queries, and ground-truth answers, offering a solid foundation for evaluating FR in complex settings.
\end{itemize}

\section{Related Work}

\subsection{Federated Retrieval}
Federated search \cite{shokouhi2011federated}, extended into RAG by combining privacy-preserving federated learning (FL) \cite{zhang2021survey} with RAG \cite{lewis2020retrieval}, enables retrieval across decentralized sources without sharing raw data\cite{chakraborty2025federated}, which has seen widespread adoption in privacy-critical domains such as healthcare \cite{jiang2024clinical, jung2025federated, xiong2024benchmarking}, finance, and legal services \cite{addison2024c}. Prior work in federated search mainly targets three aspects: (i) privacy and security through secure retrieval and encryption \cite{ jeon2021privacy, peng2021differentially}, (ii) retrieval efficiency via query routing \cite{ wang2024feb4rag}, and (iii) integration of FL and RAG tailored to domain-specific tasks \cite{wang2024unims, zeng2024federated, shojaee2025federated}. 
These advances have proven effective~\cite{zhao2024frag, xu2022detrust,wang2024unims, zeng2024federated, shojaee2025federated}; however, retrieving high-quality results in complex semantic scenarios remains challenging~\cite{wang2024feb4rag}. Existing FR methods, prompt-based and embedding-based, struggle in this setting because queries often misalign with knowledge structures, which reduces dense vector accuracy \cite{huang2021efficient}, and LLM-based prompt rewriting lacks fine-grained semantic precision \cite{hyde}.


\subsection{Neural Information Flow}

Recent studies have shown that LLMs, like the human cortex~\cite{arbib2003handbook,hawrylycz2012anatomically,zador2019critique,wang2024knowledge}, exhibit functional partitioning across their architecture~\cite{dhamdhere2018important,yu2018nisp,lecun1989optimal}.
Such functional partitions may emerge in the form of attention heads~\cite{zheng2024attention,yin2025attention,wu2024retrieval}, feed-forward networks~\cite{bandarkar2024layer,wendler2024llamas,sun2025transformer}, or neurons~\cite{huo2024mmneuron,lape}, which are shaped during training and contribute differently across tasks~\cite{dhamdhere2018important,yu2018nisp}.
Building on this partitions, information dynamically flows among these functional modules, forming a Dynamic Information Flow (DIF)~\cite{stolfo2023mechanistic,yu2024interpreting}
To leverage DIF for downstream tasks, researchers attempt to detect DIF by quantifying the attribution of parameters in the LLM with respect to input query.
Quantitative attribution methods have been extensively explored using various techniques, including forward-based methods~\cite{lianglocate,todd2023function,jiang2025unlockingpowerfunctionvectors,dai2021knowledge} and backward-based methods~\cite{feng2025recurrentknowledgeidentificationfusion,feng2024tasl} or their combinations~\cite{xu2024parenting}.
Among these methods, Shapley value-based approaches~\cite{ghorbani2020neuron,adamczewski2024shapleypruningneuralnetwork} have gained wide adoption. 
Although there has been extensive research on DIFs in large models, leveraging their powerful capabilities for modeling user intent and knowledge bases remains largely unexplored.


\begin{figure*}[t]
  \centering
\includegraphics[width=0.91\textwidth]{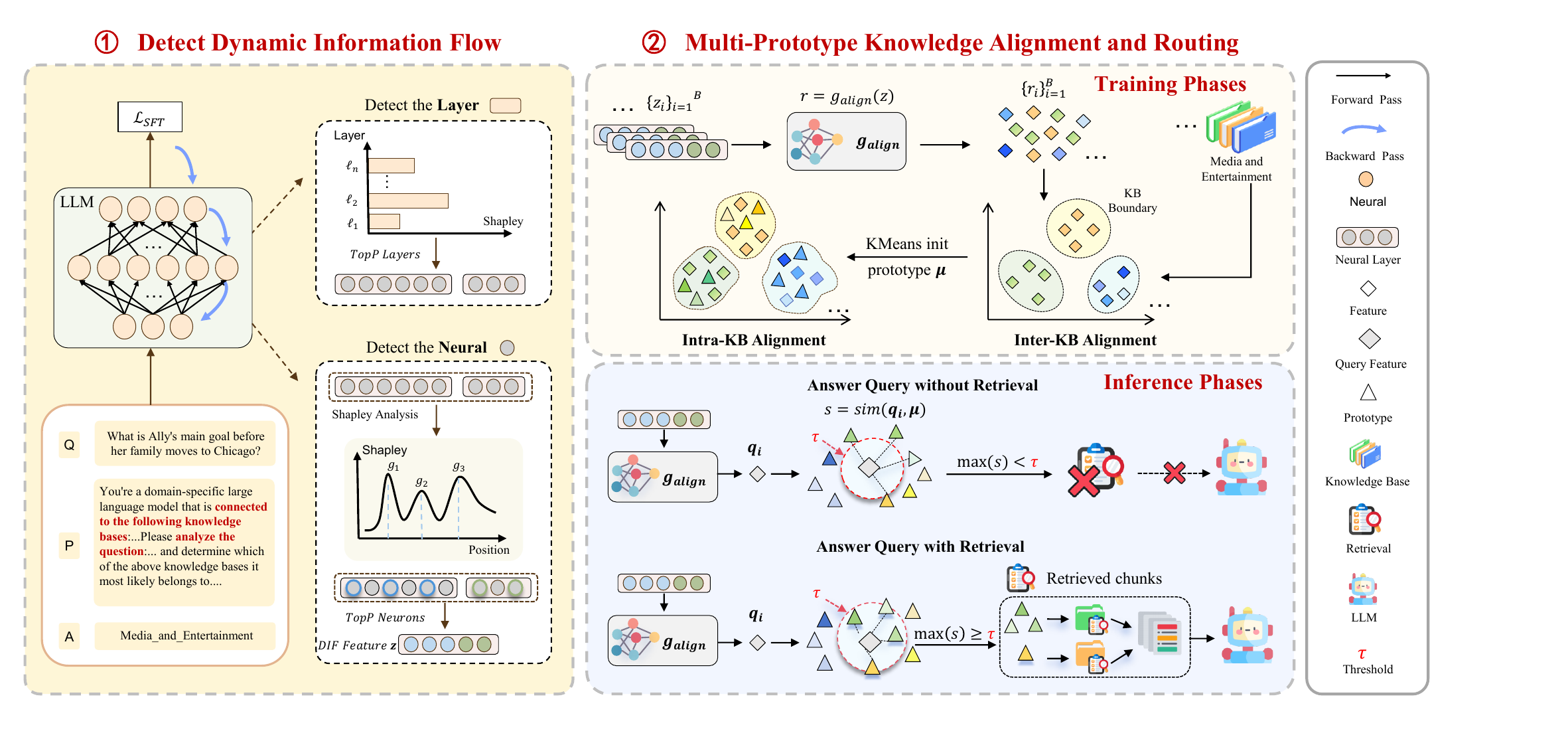}
\captionsetup{font=footnotesize}
  \caption{
  DFAMS dynamically detects relevant information flow in LLMs and employs multi-prototype alignment and routing to accurately associate queries with domain-specific knowledge bases.
  }
  \label{fig:framework}
  \vspace{-0.4cm}
\end{figure*}

\section{Method}
We introduce DFAMS, a framework designed to enable LLM to perform semantically grounded FR across distributed knowledge bases, as illustrated in Figure~\ref{fig:framework}. DFAMS operates in three stages: (1) formalizing the FR setting (Section~\ref{method:ProblemDefine}); (2) extracting query-specific internal representations through {Dynamic Information Flow} (DIF) modeling (Section~\ref{method:DynamicNeuron}); and (3) aligning these representations with structured knowledge prototypes for adaptive, multi-source routing (Section~\ref{method:Multi-Prototype}). All key notations are summarized in 
Appendix A.

\subsection{Problem Definition} 
\label{method:ProblemDefine}
\paragraph{Federated Retrieval Formalization.}
We formulate FR as a distributed retrieval problem over $I$ isolated data sources, where each source $i \in \{1, \dots, I\}$ privately hosts a knowledge base $\mathcal{K}_i = \{d_{i\ell}\}_{\ell=1}^{M_i}$ with $M_i$ documents. Due to strict privacy constraints, sources cannot exchange raw documents or intermediate representations. Given a user query $x$, the system must determine a routing vector:
\begin{equation}
f_{\text{route}}: x \mapsto \mathbf{w} = [w_1, w_2, \dots, w_I], \quad w_j \in \mathbb{N}_0, \notag
\end{equation}
where $w_j$ specifies how many documents to retrieve from $\mathcal{K}_j$. The challenge lies in selecting the most relevant sources adaptively while avoiding unnecessary retrieval.

\paragraph{RAG in FR.}
Following RAGRoute~\cite{guerraoui2025efficient}, we combine two types of knowledge: (\textit{i}) parameterized knowledge $\Theta$ stored in the model weights, and (\textit{ii}) non-parameterized knowledge $\mathcal{D} = \{\mathcal{K}_1, \dots, \mathcal{K}_I\}$, representing distributed, domain-specific corpora. Given a query $x$, the goal is to generate a reliable response: 
$\texttt{Response} \leftarrow \Theta(x, R \mid \mathcal{P})$,
where $\mathcal{P}$ is a task-specific prompt, and $R$ is the subset of knowledge bases deemed relevant. 
To handle queries answerable without external retrieval, we include an \texttt{Others} category solved solely by $\Theta$~\cite{su2024dragin}.
During training, supervision is single-source: each $(x, \mathcal{K}_i, y)$ is paired with a single knowledge base or labeled as \texttt{Others}. At inference, the model generalizes to \textit{no-source}, \textit{single-source}, or \textit{multi-source} retrieval by predicting per-source relevance scores and selecting those above a dynamic threshold $\delta$.

\subsection{Dynamic Information Flow Modeling}
\label{method:DynamicNeuron}
To accurately interpret user intent and uncover domain-relevant semantics for routing (\textbf{C1}), we detect 
DIF
capturing which neurons contribute most to domain-sensitive behavior, producing a compact, semantically grounded representation.

The process consists of two steps: (1) constructing a controlled probing dataset to isolate domain-selection behaviors; (2) identifying key layers and neurons via gradient-based Shapley attribution, and aggregating their activations into a DIF embedding.

\paragraph{Probing Dataset Construction.}
Using benchmark training or test sets for attribution often yields spurious signals, as they contain mixed user specific intents. 
To address this, we construct a dedicated probing dataset 
$\mathcal{D}_{\text{probe}} = \{(x_i, \mathcal{K}_i)\}_{i=1}^{n_{\text{probe}}}$,
where each query $x_i$ is synthesized with fixed instruction-style prompts explicitly designed to elicit domain-selection (e.g., asking the model to identify the most relevant knowledge base). Each label $\mathcal{K}_i \in \{1, \dots, I\}$ specifies the correct knowledge base. This dataset is disjoint from $\mathcal{D}_{\text{train}}$ and $\mathcal{D}_{\text{test}}$ to ensure no leakage. 

\paragraph{Neuron Attribution and DIF Embedding.}
Using $\mathcal{D}_{\text{probe}}$, we estimate the importance of each neuron using Shapley-based attribution. For a transformer block at layer $t$, the feed-forward network (FFN) computes each neuron activation as:
$\theta_{t,j} = \textsc{Act}([h_t W_{t1} + b_{t1}]_j)$,
where $h_t \in \mathbb{R}^d$ is the output of the attention sublayer, $W_{t1} \in \mathbb{R}^{d \times 4d}$ and $b_{t1} \in \mathbb{R}^{4d}$ are projection weights and biases, and $\textsc{Act}(\cdot)$ denotes the nonlinearity.
For each parameter $\theta_j \in \Theta$, its Shapley value $\phi_j$ is:
\begin{footnotesize}
    \begin{equation}
    \label{eq:shapley}
    \phi_j = -g_j^{(\gamma)} \theta_j - \frac{1}{2} \omega_{jj}^{(j)} \theta_j^2 H_{jj}^{(\gamma)} 
    - \frac{1}{2} \theta_j \sum_{k \ne j} \omega_{jk}^{(\mathcal{S})} H_{jk}^{(\gamma)} \theta_k,
    \end{equation}
\end{footnotesize}
where $g_j^{(\gamma)} = \partial \mathcal{L}_{\text{SFT}}/\partial \theta_j$ is the supervised loss gradient, and $H_{jk}^{(\gamma)}$ the Hessian for second-order interactions. The coefficients $\omega_{jj}^{(j)}$ and $\omega_{jk}^{(\mathcal{S})}$ weight self and pairwise contributions, respectively.


We then:  
\ding{182}. Select the top layers $\mathcal{L}_{\text{top}}$ based on aggregated Shapley scores.  
\ding{183}. Within each selected layer $\ell \in \mathcal{L}_{\text{top}}$, identify the most informative neuron groups $\mathcal{G}_\ell$ (adjacent units with the highest $\phi_j$).  
\ding{184}. For a query $x$, collect the activations of these groups and concatenate them to form the DIF representation:
$\mathbf{z}(x) = \textsc{Concat}\Big( \big\{ h_{\ell}^{(g)}(x) \;\big|\; g \in \mathcal{G}_\ell \big\}_{\ell \in \mathcal{L}_{\text{top}}} \Big)$ captures both semantic intent and domain cues, and serves as the input for both inter- and intra-knowledge-base modeling (Section~\ref{method:Multi-Prototype}).

\subsection{Multi-Prototype Knowledge Alignment \& Routing}
\label{method:Multi-Prototype}
To bridge internal DIF representations and external distributed knowledge structure (\textbf{C2}), we map $\mathbf{z}$ to a semantic space using a projection $g_{\text{align}}$, producing $\mathbf{r} = g_{\text{align}}(\mathbf{z})$. We train it with two contrastive stages and use it for prototype-guided routing.


\paragraph{Inter-KB Alignment.}  
We apply supervised contrastive learning to model the boundary between different knowledge bases:
\begin{footnotesize}
\begin{equation}
    \mathcal{L}_{\text{CL}} = -\sum_{i} \frac{1}{|P(i)|} \sum_{p \in P(i)} 
\log \frac{\exp(\mathbf{r}_i^\top \mathbf{r}_p/\tau_{cl})}
{\sum_{a \in A(i)} \exp(\mathbf{r}_i^\top \mathbf{r}_a/\tau_{cl})}, 
\end{equation}
\end{footnotesize}
where $P(i)$ denotes all in-batch positive samples that share the same knowledge base as $ i $, excluding $ i$ itself, and $ A(i)$ includes all other in-batch samples except $ i $. $\tau \in \mathbb{R}^+$ is a scalar temperature parameter.

\paragraph{Intra-KB Alignment.}  
Besides inter-KB modeling, we perform intra-KB modeling to capture fine-grained variations. Specifically, we first cluster embeddings within each class into prototypes $\{\boldsymbol{\mu}_m\}_{m=1}^M$ using KMeans to obtain initial cluster centers, which are then used to initialize and optimize $\mathcal{L}_{\text{PCL}} $for more detailed fine-grained modeling:
\begin{footnotesize}
    \begin{equation}
        \mathcal{L}_{\text{PCL}} = -\sum_i \frac{1}{|C(i)|} \sum_{m \in C(i)} 
    \log \frac{\exp(\mathbf{r}_i^\top \boldsymbol{\mu}_m/\tau_{pcl})}
    {\sum_{j \in AC(i)} \exp(\mathbf{r}_i^\top \boldsymbol{\mu}_j/\tau_{pcl})}, 
    \end{equation}
\end{footnotesize}

where $C(i)$ denotes the set of prototypes most similar to $\mathbf{r}_i$ based on cosine similarity (typically the nearest prototype), and $AC(i)$ represents all prototypes excluding those in $C(i)$.
During inference, given a query embedding $\mathbf{q}$, we compute its similarity scores $s_i = \text{sim}(\mathbf{q}, \boldsymbol{\mu}_i)$ to all learned prototypes $\boldsymbol{\mu}_i$. The routing function then proceeds in two stages:

\paragraph{Adaptive Triggering.} If the maximum similarity falls below a threshold $\tau$, i.e., $\max_i s_i < \tau$, the system abstains from retrieval.
\begin{equation}
    f_{\text{route}}(\mathbf{q}) = 
\begin{cases}
\mathbf{0}, & \max_i s_i < \tau,\\
[w_1, \dots, w_K], & \text{otherwise}
\end{cases}
\end{equation}

\paragraph{Semantic Routing.} Otherwise, it identifies the top-$N$ most similar prototypes $\mathcal{I}$ and allocates a total of $T$ retrieval slots across knowledge bases. The number of documents assigned to each knowledge base $k$ is computed as:
\begin{equation}
    w_k = \left\lfloor \frac{\sum_{i \in \mathcal{I},\,k_i=k} s_i}
{\sum_{k'} \sum_{i \in \mathcal{I},\,k_i=k'} s_i} \cdot T \right\rfloor
\end{equation}
This two-step mechanism ensures retrieval occurs only when necessary, with resources allocated by prototype-level semantic relevance.

\begin{table*}[ht]
\centering

\fontsize{9pt}{9pt}\selectfont
\setlength{\tabcolsep}{4pt} 
\renewcommand{\arraystretch}{1.2}
\resizebox{\linewidth}{!}{ 
\begin{tabular}{l|ccc|ccc|cc|c|c}
\toprule

\textbf{Method} 
& \multicolumn{8}{c|}{\textbf{In-Domain}} 
& \multicolumn{2}{c}{\textbf{Out-of-Domain}} \\
\cmidrule(lr){2-9} \cmidrule(lr){10-11}

& \multicolumn{3}{c|}{Wiki} & \multicolumn{3}{c|}{Med} 
& \multicolumn{2}{c|}{PEP} & MMLU & MIRAGE \\
& \textbf{Cls Acc ($\uparrow$)} & \textbf{Recall ($\uparrow$)} & \textbf{QA Acc ($\uparrow$)} 
& \textbf{Cls Acc ($\uparrow$)} & \textbf{Recall ($\uparrow$)} & \textbf{QA Acc ($\uparrow$)} 
& \textbf{Cls Acc ($\uparrow$)} & \textbf{QA Score ($\uparrow$)} 
& \textbf{QA Acc ($\uparrow$)} & \textbf{QA Acc ($\uparrow$)} \\
\midrule

\multicolumn{11}{c}{\textit{Qwen2.5-7B}} \\
\midrule
No RAG        & /      & /      & 59.30  & /     & /      & 75.20  & /      & 7.56  & 79.05 & 68.05 \\
Merged-RAG    & /      & 48.49  & 77.20  & /     & 33.89  & 79.60  & /      & 8.33  & 77.74 & 65.78 \\
Prompt        & 32.47  & 25.93  & 63.66  & 48.82 & 34.68  & 79.59  & 66.86  & 6.21  & 80.77 & 69.17 \\
CoT Prompt    & 58.67  & 38.17  & 71.40  & 48.82 & 31.40  & 80.72  & 77.91  & 5.92  & 80.52 & 69.47 \\
SFT    & 41.13  & 25.60  & 61.97  & 49.35 & 24.70  & 75.73  & 68.76  & 7.23  & 78.23 & 67.97 \\
RopMura       & 62.83  & 48.92  & 77.47  & 53.09 & 41.59  & 82.01  & 75.92  & 7.18  & 80.38 & 69.97 \\
RAGRoute      & 76.07  & 50.04  & 78.40  & 69.04 & 35.78  & 73.64  & 51.47  & 7.14  & 80.20 & 69.77 \\
\textbf{DFAMS} & \textbf{85.03} & \textbf{53.83} & \textbf{78.94} & \textbf{71.81} & \textbf{42.82} & \textbf{82.82} & \textbf{82.85} & \textbf{8.39} & \textbf{86.17} & \textbf{79.88} \\
\midrule

\multicolumn{11}{c}{\textit{LLaMA3.1-8B}} \\
\midrule
No RAG        & /      & /      & 60.40  & /     & /      & 68.27  & /      & 5.35  & 59.31 & 61.90 \\
Merged-RAG    & /      & 48.49  & 76.48  & /     & 33.89  & 74.17  & /      & 5.38  & 56.30 & 62.68 \\
Prompt        & 36.18  & 25.93  & 64.67  & 49.57 & 34.72  & 71.65   & 65.99  & 5.17  & 63.90 & 61.51 \\
CoT Prompt    & 58.20  & 38.14  & 68.98  & 48.92 & 31.59  & 74.46  & 76.75  & 4.82  & 63.55 & 63.43 \\
SFT  & 42.61  & 26.12  & 63.78  &49.31  & 26.41  & 70.41  & 63.95  & 4.86  & 58.31  &62.11 \\
RopMura       & 62.83  & 48.92  & 76.78  & 53.09 & 41.59  & 75.37 & 75.92  & 4.46  & 64.06 & 61.82 \\
RAGRoute      & 76.07  & 50.04  & 76.83  & 69.04 & 35.78  & 74.55  & 51.47  & 4.58  & 64.37 & 63.16 \\
\textbf{DFAMS} & \textbf{84.36} & \textbf{53.83} & \textbf{78.60} & \textbf{72.57} & \textbf{42.82} & \textbf{77.65} & \textbf{81.98} & \textbf{8.08} & \textbf{64.89}& \textbf{64.89} \\
\bottomrule
\end{tabular}
}
\caption{Performance comparison (\%) on \textit{Wiki}, \textit{Med}, \textit{PEP}, \textit{MMLU}, and \textit{MIRAGE}, where \textbf{bold} indicates the best result, and symbol slash "/" denotes inapplicable metrics
.
}
\label{tab:comparison}
\vspace{-0.4cm}
\end{table*}

\section{Experiments}
We conduct extensive experiments across multiple datasets to evaluate the effectiveness of DFAMS in Federated Retrieval settings and answer the following key research questions:
\begin{itemize}[leftmargin=*,noitemsep,topsep=2pt]
    \item \textbf{RQ1 (Section~\ref{Main Results}):} 
    Does DFAMS consistently outperform existing advanced methods?
    \item \textbf{RQ2 (Section~\ref{Ablation Study}):} 
        How do the proposed component contribute to performance improvements?
    \item \textbf{RQ3 (Section~\ref{Sensitivity Analysis}):} 
How sensitive is DFAMS to variations in model configurations?
\end{itemize}

\subsection{Experimental Setup}

\paragraph{LLM Backbones.}
 We implement DFAMS on four open-weight LLMs with varying scales to evaluate scalability and generalization: \textit{Qwen2.5-0.5B}, \textit{Qwen2.5-3B}, \textit{Qwen2.5-7B}~\cite{DBLP:journals/corr/abs-2412-15115}, and \textit{Llama3.1-8B}~\cite{grattafiori2024llama}. 

\paragraph{Retrieval Configuration.}
Following \citet{wu2025talk} and \citet{guerraoui2025efficient}, DFAMS utilizes FAISS~\cite{douze2024faiss} for dense vector retrieval across three FR scenarios, each comprising multiple knowledge bases. 
For each query, top-$10$ chunks are retrieved from selected sources across all relevant knowledge bases within the same scenario. 
Further details are provided in 
Appendix~\ref{appendix:retrieval}
\paragraph{Datasets.} 
We construct three in-domain evaluation benchmarks, \textit{Wiki}, \textit{Med}, and \textit{PEP}. 
To further evaluate out-of-domain generalization, we test models trained on \textit{Wiki} and \textit{Med} using a subset of \textit{MMLU}~\cite{hendrycks2020measuring} and a subset of \textit{MIRAGE}~\cite{xiong2024benchmarking}, following the setup of ~\cite{guerraoui2025efficient}
. 
Further construction details are provided in 
Appendix~\ref{appendix:datasets}.

\paragraph{Baselines.}
We compare DFAMS with several representative retrieval and routing strategies: \textit{No-RAG}, \textit{Merged-RAG} 
, and two prompt-based methods \textit{Prompt} and \textit{CoT Prompt}~\cite{wei2023chainofthought}.
We further compare with a supervised fine-tuning baseline (\textit{SFT}) 
\cite{hu2021lora}.
In addition, we consider two recent multi-source retrieval approaches: \textit{RAGRoute}~\cite{guerraoui2025efficient}, which employs a binary classifier for each knowledge base to select the Top-$k$ sources, and \textit{RopMura}~\cite{wu2025talk}, a prototype-based multi-agent routing method.
Implementation details are provided in Appendix~\ref{appendix:baselines}.

\paragraph{Evaluation Metrics.} 
We report three complementary metrics: 
(1) \textbf{Cls Acc}, which measures whether the predicted knowledge base(s) match the ground truth, including correct \texttt{Others} predictions for no-retrieval cases;  
(2) \textbf{Recall}, computed over Top-10 retrieved documents for retrieval-triggering queries, measuring the proportion of gold documents successfully retrieved excluding no-retrieval cases;
(3) \textbf{QA}, the end-to-end answer accuracy, using accuracy({QA ACC}) for multiple-choice and LLM-based scoring ({QA Score})~\cite{zheng2023judging} for open-ended questions.
More metric details are available in 
Appendix~\ref{appendix:metrics}.

\subsection{Main Result Analysis (RQ1)}
\label{Main Results}
To address \textit{RQ1}, we train DFAMS on the \textit{Wiki}, \textit{Med}, and \textit{PEP} datasets, and evaluate it on both in-domain (Wiki, Med, PEP) and out-of-domain (MMLU, MIRAGE) benchmarks using \textit{Qwen2.5-7B} and \textit{LLaMA3.1-8B}, comparing against baselines from retrieval, prompt-based, and multi-source routing methods.
\paragraph{Comparison with Baselines.}  

Table~\ref{tab:comparison} summarizes the performance of DFAMS compared with a range of baseline methods across three key metrics: Cls Acc, Recall, and QA Acc/Score. DFAMS consistently outperforms all baselines on both in-domain and out-of-domain datasets.
\paragraph{Comparison of Advanced Multi-Source Retrieval and Naive Methods.}
In most cases, multi-source retrieval methods (\textit{RopMura}, \textit{RAGRoute}) achieve higher Cls Acc and Recall than \textit{Prompt}-based or \textit{Merged-RAG} baselines in most cases. For example, on the Wiki dataset with \textit{Qwen2.5-7B}, RAGRoute achieves a Cls Acc of 76.07\% and recall of 50.04\%, compared to just 32.47\% and 25.93\% from Prompt. Similarly, RopMura outperforms Merged-RAG on Med in Recall (41.59\% vs. 33.89\%). While multi-source retrieval methods (\textit{RopMura}, \textit{RAGRoute}) generally achieve higher recall through broader source coverage, these gains often come at the cost of increased cross-domain noise. For instance, although RAGRoute obtains higher recall on the Med dataset, its QA accuracy (73.64\%) falls short of that of Merged-RAG (79.60\%). In contrast, prompt-based methods adopt a more conservative source selection strategy, often retrieving from fewer knowledge bases, which helps reduce cross-domain noise.  Merged methods, on the other hand, rely on dense semantic similarity across the entire corpus;  while the retrieved chunks may not always be precisely grounded, they tend to be semantically coherent. 

\paragraph{Comparison of DFAMS and Advanced Multi-Source Retrieval.}
Compared with advanced multi-source retrieval methods, {DFAMS} consistently achieves higher Cls Acc and Recall across all datasets and backbones. For instance, on Wiki with \textit{Qwen2.5-7B}, DFAMS outperforms RAGRoute by +8.96\% in Cls Acc (85.03\% vs. 76.07\%) and +3.79\% in Recall (53.83\% vs. 50.04\%). These improvements stem from DFAMS’s fine-grained modeling of DIF, enabling more accurate identification of query intent and relevant knowledge sources.  This leads to better routing and more focused retrieval with less cross-domain noise.  As a result, DFAMS consistently achieves the highest QA accuracy across all datasets.  For example, on the \textit{Qwen2.5-7B} backbone, it reaches 82.82\% QA accuracy on Med and 78.94\% on Wiki, outperforming both RopMura and RAGRoute.
\paragraph{Adaptive Retrieval Capability.}
DFAMS learns to decide whether external retrieval is needed for a given query. We evaluate its adaptive retrieval capability by grouping all sources into a single \texttt{Knowledge} class and using an \texttt{Others} class for queries answerable via parametric knowledge. As shown in Table~\ref{tab:adaptive}, DFAMS achieves high accuracy on Wiki (99.95\%) and Med (93.67\%), closely matching \textit{Probing RAG} and significantly outperforming the \textit{Prompt}-based approach. These results highlight DFAMS’s ability to avoid unnecessary retrieval while preserving high coverage when external information is required. 
\begin{table}[ht]
\vspace{-0.1cm}
\centering
\fontsize{9pt}{9pt}\selectfont
\renewcommand{\arraystretch}{1.2}
\setlength{\tabcolsep}{6pt}
\begin{tabularx}{0.9\linewidth}{l|>{\centering\arraybackslash}X>{\centering\arraybackslash}X}
\toprule
\rowcolor{gray!10}
\textbf{Method} & \textbf{Wiki} Acc ($\uparrow$) & \textbf{Med} Acc ($\uparrow$) \\
\midrule
Prompt & 69.73 & 84.20 \\
Probing RAG & \textbf{99.98} & 92.80 \\
\textbf{DFAMS} & 99.95 & \textbf{93.67} \\
\bottomrule
\end{tabularx}

\caption{Adaptive retrieval accuracy comparison between different methods on Wiki and Med datasets.}
\label{tab:adaptive}
\vspace{-0.3cm}
\end{table}



\paragraph{Inference Efficiency.}  
We assess DFAMS’s efficiency by measuring average routing, retrieval, and total latency per sample on the Med dataset (Table~\ref{tab:time-cost}). Despite relying on LLMs, DFAMS achieves a low end-to-end latency of 1.48s—substantially faster than \textit{Prompt} (15.59s) and \textit{Merged-RAG} (3.89s). Compared to \textit{RAGRoute} (1.64s), DFAMS is slightly faster due to retrieving from fewer sources. In RAGRoute, its pursuit of higher recall often triggers more knowledge bases, and slower sources can increase latency despite parallel execution. Overall, DFAMS offers faster inference with precise routing.
 \begin{table}[ht]
\centering
\fontsize{9pt}{9pt}\selectfont
\setlength{\tabcolsep}{6pt}
\renewcommand{\arraystretch}{1.2}
\begin{tabular}{l|ccc}
\toprule
\rowcolor{gray!10}
\textbf{Method} & \textbf{Routing (s)} & \textbf{Retrieval (s)} & \textbf{Total (s)} \\
\midrule
Prompt & 14.25 & 1.35 & 15.59 \\
Merged-RAG & \textbf{0} & 3.89 & 3.89 \\
RAGRoute & 0.0023 & 1.64 & 1.64 \\
\textbf{DFAMS} & 0.13 & \textbf{1.34} & \textbf{1.48} \\
\bottomrule
\end{tabular}
\caption{Comparison of routing, retrieval, and total processing times for different methods on the Med dataset.}
\label{tab:time-cost}
\vspace{-0.4cm}
\end{table}

\subsection{Ablation Study (RQ2)}
\label{Ablation Study}
To answer \textbf{RQ2}, we conduct ablation studies on the two core components of DFAMS: (1) \textit{Dynamic Information Flow Modeling} and (2) \textit{Multi-Prototype Knowledge Alignment and Routing}, aiming to investigate their individual contributions to the overall system performance.

\textbf{Dynamic Information Flow Modeling.}  
We conduct ablation studies to test our central hypothesis: that DIF signals not only exist but can be effectively detected and exploited to improve model performance. We design two experimental settings:
(1) Frozen LLM w/o Align-MLP: We remove the trainable \texttt{Align-MLP} and directly utilize the extracted DIF. This setting investigates whether native LLM activations inherently encode subdomain-aware signals—i.e., whether meaningful associations between user queries and knowledge subdomains can be inferred without explicit alignment.
(2) Full DFAMS (w/ Align-MLP): We enable the trainable \texttt{Align-MLP} to utilize DIF to assesse whether modeling the knowledge base on DIF leads to improved alignment and enhanced downstream performance.

\begin{table}[ht]
\vspace{-0.1cm}
\centering
\fontsize{9pt}{9pt}\selectfont
\renewcommand{\arraystretch}{1.2}
\setlength{\tabcolsep}{6pt}
\begin{tabularx}{0.75\linewidth}{l|>{\centering\arraybackslash}X>{\centering\arraybackslash}X}
\toprule
\rowcolor{gray!10}
\textbf{Method} & \textbf{Wiki} ($\uparrow$) & \textbf{Med} ($\uparrow$) \\
\midrule
\multicolumn{3}{c}{\textit{Frozen (no \texttt{Align-MLP})}} \\
\midrule
Random & 53.19 & 42.90 \\
Full & 64.05 & 52.67 \\
\textbf{DFAMS} & \textbf{67.74} & \textbf{54.68} \\
\midrule
\multicolumn{3}{c}{\textit{Trained (with \texttt{Align-MLP})}} \\
\midrule
Random & 81.49 & 66.47 \\
Full & 82.42 & 68.36 \\
\textbf{DFAMS} & \textbf{85.03} & \textbf{71.81} \\
\bottomrule
\end{tabularx}
\caption{Ablation analysis of Dynamic Information Flow modeling on Wiki and Med datasets}
\label{tab:flow-ablation}
\vspace{-0.3cm}
\end{table}

We conduct experiments on both the \textit{Wiki} and \textit{Med} datasets. 
Table~\ref{tab:flow-ablation} shows that in setting (1), DFAMS (2000-dim) outperforms both random (2000-dim) and full-layer (3584-dim) baselines, achieving +14.6\% accuracy gain on Wiki and +11.78\% on Med over the random baseline. These results validate our hypothesis: DIF captures query-subdomain associations and can be directly leveraged, even without further training. 
Moreover, DFAMS also surpasses the Full baseline, yielding +3.69\% and +2.01\% accuracy improvements on the Wiki and Med datasets, respectively.
In setting (2), with \texttt{Align-MLP} enabled, DFAMS achieves +3.54\% and +2.61\% higher accuracy than the random and full-layer baselines on Wiki, and +5.34\% and +3.45\% on Med. These results highlight the benefit of leveraging DIF for knowledge base modeling, resulting in better alignment and improved downstream performance. Additional results and analyses are provided in 
Appendix~\ref{appendix:layer-selection}.
\begin{table}[ht]
\vspace{-0.1cm}
\centering
\fontsize{9pt}{9pt}\selectfont
\setlength{\tabcolsep}{6pt}
\renewcommand{\arraystretch}{1.1}
\begin{tabular}{l|c|c}
\toprule
\rowcolor{gray!10}
\textbf{Variant} & \textbf{Accuracy} ($\uparrow$) & \textbf{Recall} ($\uparrow$) \\
\midrule
\textbf{Full Method} & \textbf{85.03} & \textbf{53.83} \\
\midrule
\textit{w/o Inter-KB Alignment} & 75.89 & 52.09 \\
\textit{w/o Intra-KB Alignment} & 83.28 & 50.68 \\
\textit{w/o Adaptive Triggering} & 79.56 & \textbf{53.83} \\
\textit{w/o Semantic Routing} & 80.67 & 48.67 \\
\bottomrule
\end{tabular}
\caption{Ablation Analysis of Multi-Prototype Alignment and Routing Components on Wiki dataset}
\label{tab:model_method_comparison}
\vspace{-0.3cm}
\end{table}

\textbf{Ablation on Multi-Prototype Knowledge Alignment and Routing}
We conduct targeted ablations to evaluate the impact of DFAMS’s core components. Results are shown in Table~\ref{tab:model_method_comparison}.
Disabling {inter-KB alignment}, which separates semantic boundaries across knowledge bases, causes the Cls ACC drop (-9.14\%), highlighting its key role in knowledge base selection and adaptive retrival. Removing {intra-KB alignment}, responsible for modeling knowledge bases' subdomain structures via multi-prototype contrastive learning, leads to the biggest recall decline (-3.15\%), showing its importance for accuracy and high  quality retrieval. On the inference side, removing {adaptive triggering}  reduces accuracy (-5.47\%), as the model can no longer skip unnecessary retrieval. Disabling {Semantic Routing}, which confines retrieval to only one top source, further decreases recall (-5.16\%), highlighting the value of semantic-aware resource allocation across multiple knowledge bases. 



\subsection{Sensitivity Analysis (RQ3)}
\label{Sensitivity Analysis}
\begin{figure}[htb]
\vspace{-0.1cm}
  \centering
  \includegraphics[scale=0.18]{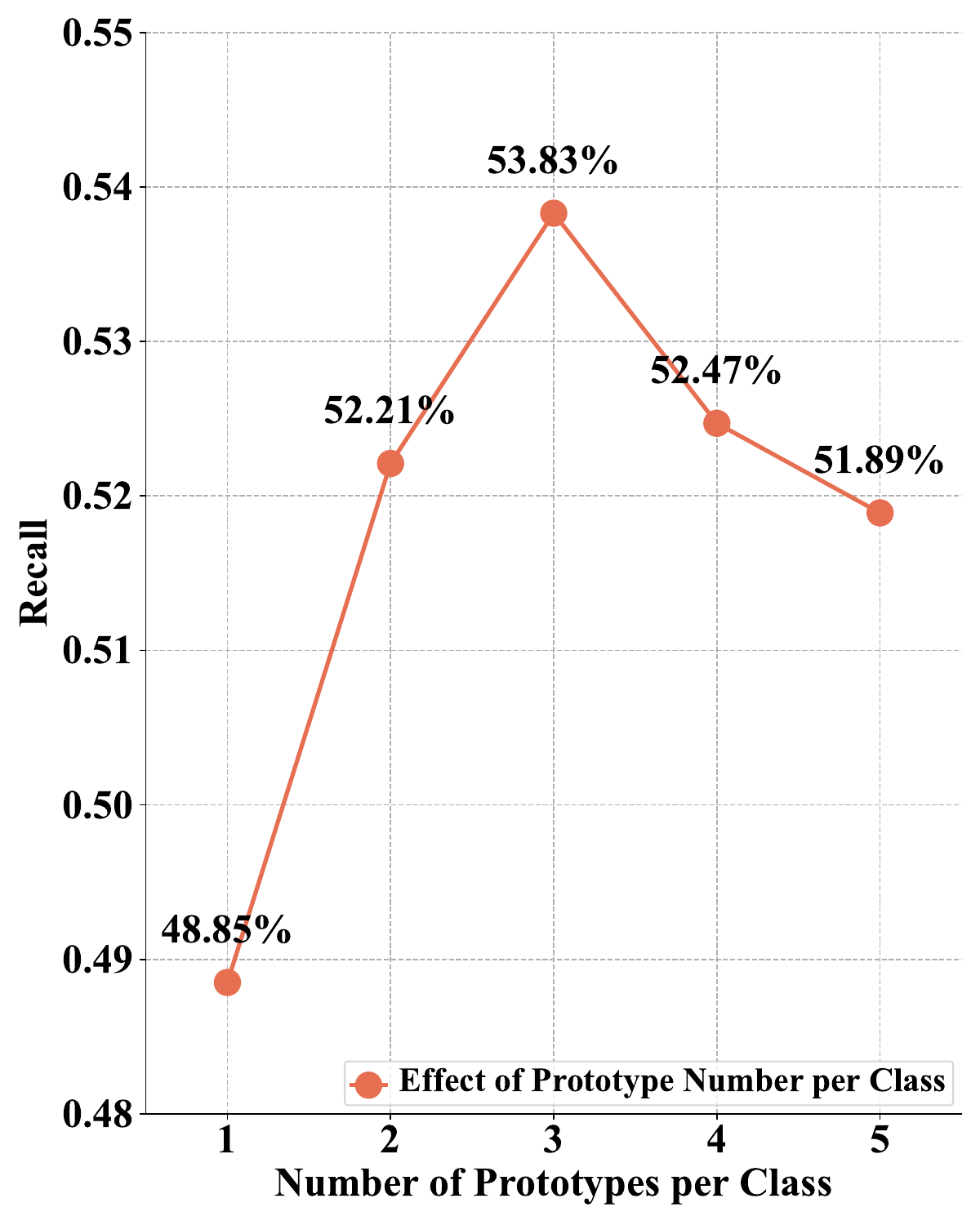}
  \includegraphics[scale=0.18]{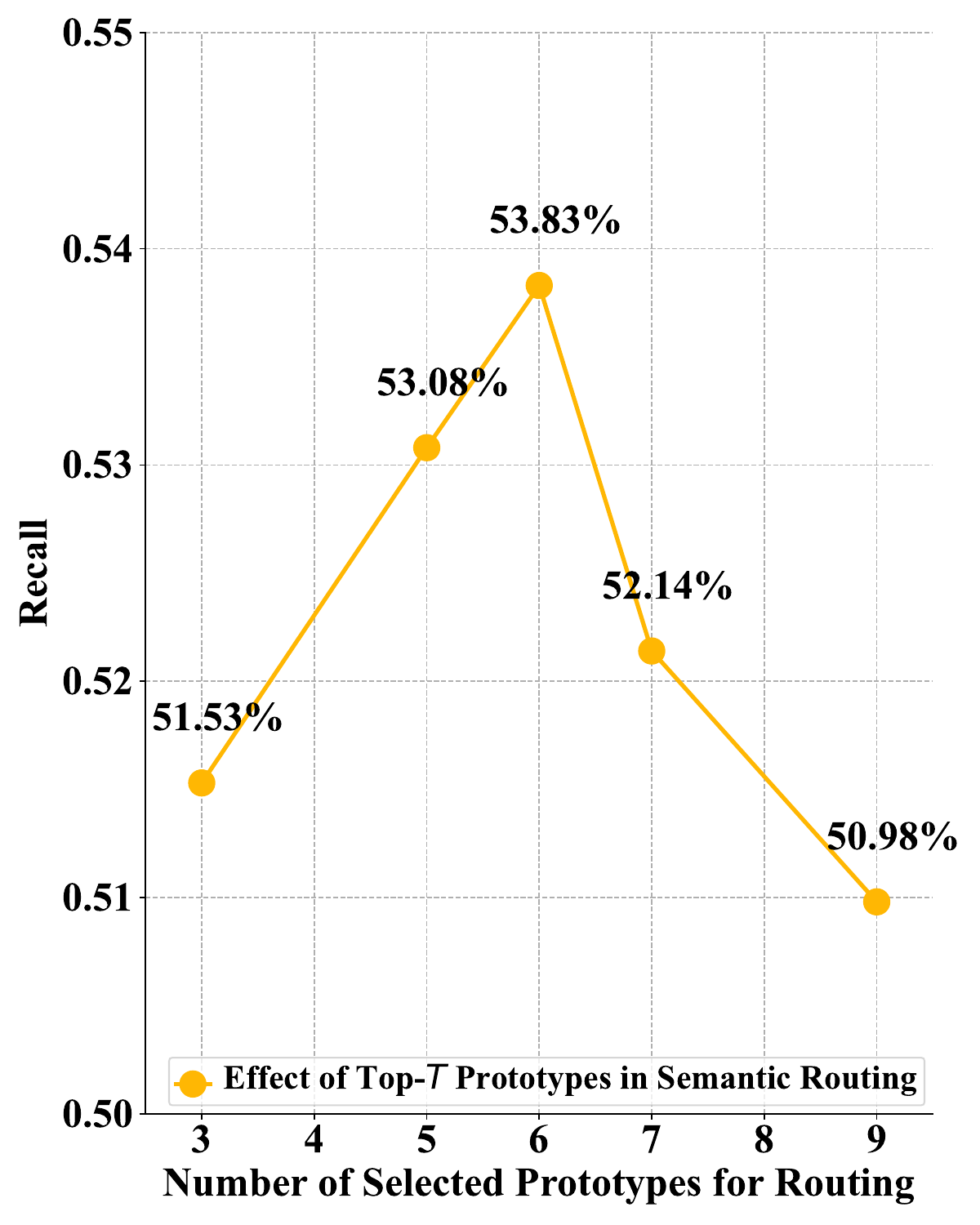}
  \captionsetup{font=footnotesize}
  \caption{Hyperparameter Analysis of Prototypes per Class (Left) and Selected Prototypes for Routing (Right)}
   \label{fig:proto_sensitivity}
\vspace{-0.3cm}
\end{figure}
To assess how sensitive DFAMS is to variations in key configurations, we conduct experiments on the Wiki dataset.
\paragraph{Effect of Prototype Number.}
From the result in Figure~\ref{fig:proto_sensitivity} (left), the accuracy increases with prototype count, peaking at 3 (53.8\%), then declines (e.g., 52.1\% at 4), suggesting that too few prototypes underfit subdomain diversity, while too many cause over-fragmentation.
\paragraph{Effect of Semantic Routing Top-$T$.}
As shown in Figure~\ref{fig:proto_sensitivity} (right), when top-$T$ is 3, it achieves the best balance (84.2\% accuracy, 53.8\% recall). Lower $T$ values miss relevant sources, while higher $T$ dilutes document allocation across knowledge bases, reducing 
key information retrieval.

\section{Conclusion}



We propose DFAMS, a novel FR framework that explicitly models the DIF within LLMs to enhance query understanding and cross-source knowledge alignment. 
By leveraging gradient-based neuron attribution and Shapley value estimation, DFAMS identifies latent neural activation paths that reflect user intent and subdomain relevance.
The framework also introduces a Multi-Prototype Knowledge Alignment and Routing strategy, which enables fine-grained modeling of individual knowledge bases.
Our experiments show that DFAMS consistently outperforms existing FR methods 
, confirming its effectiveness in resolving semantic ambiguities and improving cross-source routing in complex settings.


\section*{Limitations}

While DFAMS achieves consistent improvements in federated retrieval, several limitations still remain. 
First of all, the framework relies on extracting Dynamic Information Flow (DIF) signals from large transformer-based LLMs; while we demonstrate scalability to smaller models, the applicability of DIF-based modeling to fundamentally different architectures has not been thoroughly examined. Second, our evaluation is conducted on curated benchmarks with static knowledge bases. In real-world applications, knowledge sources often evolve continuously; when significant updates occur, the framework may require retraining or integration with continual learning strategies, which has not yet been fully explored. Addressing these aspects could further enhance the robustness and applicability of DFAMS in broader federated retrieval scenarios.

Future efforts will focus on optimizing prototype selection and update strategies, integrating DFAMS with advanced RAG techniques for improved end-to-end reasoning, and exploring its applicability in other specialized domains such as legal, finance, and scientific research.

\section*{Ethical considerations}
To evaluate the efficacy of our work, we conducted experiments using five datasets: Wiki, Med, PEP, MMLU, and MIRAGE.   Except for PEP, all datasets are publicly available and used in accordance with their respective licenses and terms of use.   PEP was obtained and used with proper authorization.   The datasets do not contain personally identifiable information, and no human or animal subjects were directly involved in this research.

\bibliography{custom}

\appendix

\newpage
\section{Notations Table}
\label{sec:app notation}
This section presents a comprehensive list of key notations and symbols employed in the DFAMS framework.
\begin{table}[htbp]
\centering
\resizebox{1\linewidth}{!}{
\begin{tabular}{p{2.8cm} p{6.8cm}}  
\toprule
\textbf{Symbol} & \textbf{Description} \\
\midrule
$I$ & Number of isolated knowledge bases (KBs) \\
$\mathcal{K}i = {\{d_{i\ell}\}}_{\ell=1}^{M_i}$ & The $i$-th KB containing $M_i$ documents $d_{i\ell}$\\
$x$ & Input query for KBs selection and retrieval \\
$f_{\text{route}}$ & Maps $x$ to document allocations over KBs \\
$\mathbf{w} = [w_1, \dots, w_I]$ & Document allocation vector; $w_j$ is the number retrieved from $\mathcal{K}_j$ \\
$\Theta$ & Parameterized knowledge in the LLM \\
{\small $\mathcal{D} = \{\mathcal{K}_1, \dots, \mathcal{K}_I\}$ } & KBs representing non-parameterized knowledge \\
$R$ & Retrieved subset of KBs for answering query \\
$\mathcal{P}$ & Task-specific prompt \\
$y$ & Ground-truth answer to query 
$x$ \\
$\mathcal{D}_{\text{probe}}$ & Dedicated probing dataset for neuron attribution and domain selection analysis \\
$h_t$ & Output of attention sublayer in layer $t$ \\
$W_{t1}, b_{t1}$ & FFN projection weights and biases at layer $t$ \\
$\textsc{Act}(\cdot)$ & Activation function \\
$\phi_j$ & Shapley Shapley value for neuron $j$ \\
$g_j^{(\gamma)}$ & Gradient of supervised loss w.r.t.\ $\theta_j$ \\
$H_{jk}^{(\gamma)}$ & Hessian of loss capturing 2nd-order interactions \\
$\omega_{jj}^{(j)}$ & Weighting coefficient for self-contribution of neuron $j$ in Shapley approximation \\

$\omega_{jk}^{(\mathcal{S})}$ & Weighting coefficient for pairwise contribution between neurons $j$ and $k$ \\
$\mathbf{z}$ & DIF embedding composed of high-attribution neuron activations \\
$\textsc{Concat}(\cdot)$ & Concatenates input set into a single vector \\
\hline
$g_{\text{align}}$ & Projection function mapping $\mathbf{z}$ to aligned query embedding $\mathbf{r}$ in the semantic space \\
$\mathcal{L}_{\text{CL}}$ & supervised contrastive loss \\
$B$ & Batch size used for contrastive training \\
$P(i)$ & Set of in-batch positive samples sharing the same KB label as sample $i$ \\
$A(i)$ & All other in-batch samples excluding $i$ (i.e., positive + negative candidates) \\
$\tau_{cl}$ & Temperature scaling factor in contrastive loss\\
$\boldsymbol{\mu}_m$ & Prototype vector in the contrastive-aligned space \\
$\mathcal{L}_{\text{PCL}}$ & Intra-KB prototype contrastive loss \\
$C(i)$ & Closest prototype(s) to sample $i$\\
$AC(i)$ & All prototypes excluding those in $C(i)$ \\
$\tau_{pcl}$ & Temperature scaling factor in $\mathcal{L}_{\text{PCL}}$ \\
\hline
$\mathbf{q}$ & Embedded representation of an unseen query \\
$\text{sim}(\cdot, \cdot)$ & Cosine similarity  between two embeddings \\
$s_i$ & Similarity between $\mathbf{q}$ and prototype $\mathbf{p}_i$ \\
$T$ & Total retrieval slots to be allocated \\
$\tau$ &  Adaptive triggering threshold \\
$\mathcal{I}$ & Top-$N$ nearest prototypes \\
\bottomrule
\end{tabular}}
\caption{Key Notations used in the DFAMS framework}
\label{tab:symbols}
\end{table}
\section{Algorithm}
\label{sec: algorithm} 


In this section, we detail the full DFAMS workflow, spanning its probing, training, and inference stages. Algorithm~\ref{alg:struct_probe} identifies domain-sensitive neurons in pretrained LLMs, 
while Algorithm~\ref{alg:dfams_train} describes the training procedure. 
Finally, Algorithm~\ref{alg:dfams_infer} presents the adaptive prototype-guided routing mechanism used during inference to dynamically allocate retrieval resources based on semantic relevance.

\begin{algorithm}[H]
    \caption{Neuron Probing for DIF Extraction}
    \label{alg:struct_probe}
    \begin{algorithmic}[1]
    \Require Probing dataset $\mathcal{D}_{\text{probe}} = \{(x_i, \mathcal{K}_i)\}$, pretrained LLM $\Theta$, layer count $L$, top layer number $T$, neuron group size $G$
    \Ensure DIF-relevant neuron set $\mathcal{N}$
    
    \State Initialize Shapley values $\Phi \leftarrow \mathbf{0}$ for all neurons in $\Theta$
    
    \For{$(x_i, \mathcal{K}_i) \in \mathcal{D}_{\text{probe}}$}
        \State Compute loss $\mathcal{L}_{\text{SFT}}$ on $\Theta(x_i)$
        \State Backpropagate gradients $g_j = \frac{\partial \mathcal{L}_{\text{SFT}}}{\partial \theta_j}$
        \State Compute second-order Hessian approximations $H_{jk}$
        
        \For{$t = 1$ to $L$}
            \For{neuron $j$ in layer $t$}
                \State Compute Shapley value $\phi_j$ using Equation \ref{eq:shapley}
                \State $\Phi_{t,j} \leftarrow \Phi_{t,j} + \phi_j$
            \EndFor
        \EndFor
    \EndFor
    
    \State Average $\Phi$ over samples
    \State Select top layers $\mathcal{L}_{\text{top}}$ with highest total Shapley mass
    \State For each layer $\ell \in \mathcal{L}_{\text{top}}$, select top neuron groups $\mathcal{G}_\ell$ of size $G$
    \State Construct $\mathcal{N} = \{\, h_{\ell}^{(g)} \mid \ell \in \mathcal{L}_{\text{top}},\, g \in \mathcal{G}_\ell \,\}$
    \State \Return $\mathcal{N}$

    \end{algorithmic}
\end{algorithm}

\subsection{Neuron Probing for DIF Extraction}

To identify domain-sensitive neurons within pretrained LLMs, Algorithm~\ref{alg:struct_probe} introduces how to locate the subset most relevant for DIF. 
The process initiates by iterating over a probing dataset, capturing the model's loss and computing per-neuron gradients for each data sample. To measure each neuron's importance regarding domain sensitivity, Shapley values are approximated for all neurons across the LLM’s layers. This approach quantifies each neuron’s marginal contribution to the model’s ability to capture domain-specific information. The cumulative Shapley scores are averaged over all samples in the probing set, after which the algorithm selects the layers with the highest aggregate Shapley mass. Within these layers, the most influential neuron groups are further identified based on group-wise Shapley values. The final output is a compact set of domain-relevant neurons, $\mathcal{N}$, which serve as the basis for extracting DIF representations in subsequent stages.

\begin{algorithm}[H]
\caption{Two-Stage Contrastive Alignment of DIF}
\label{alg:dfams_train}
\begin{algorithmic}[1]
\Require Training set $\mathcal{D}_{\text{train}} = \{(x_i, y_i, \mathcal{K}_i)\}$, pretrained LLM $\Theta$, DIF neuron set $\mathcal{N}$, temperature $\tau_{cl}$, $\tau_{pcl}$, epochs $E_1, E_2$, weighting factor $\lambda$
\Ensure Optimized $g_{align}$ and prototype $\mu$

\State \textbf{Stage 0: DIF Embedding Extraction}
\State Initialize embedding set $\mathcal{Z} \leftarrow \emptyset$
\For{each $(x_i, y_i, \mathcal{K}_i) \in \mathcal{D}_{\text{train}}$}
    \State Extract DIF embedding: $\mathbf{z}_i \leftarrow \textsc{Probe}(\Theta(x_i), \mathcal{N})$
    \State Store ${\mathbf{z}}_i$ and metadata in $\mathcal{Z} \leftarrow \mathcal{Z} \cup \{({\mathbf{z}}_i, y_i, \mathcal{K}_i)\}$
\EndFor

\State \textbf{Stage 1: Inter-KB Alignment}
\For{epoch $e = 1$ to $E_1$}
    \For{each minibatch $\{(\mathbf{z}_i, y_i, \mathcal{K}_i)\}_{i=1}^{B}$}
        \State Compute aligned embeddings $\mathbf{r}_i = g_{{align}}(\mathbf{z}_i)$
        \State Compute $\mathcal{L}_{\text{cl}}$
        \State Update $g_{align}$ parameters via backpropagation
    \EndFor
\EndFor

\State \textbf{Stage 2: Intra-KB Alignment}
\State Compute aligned embeddings $\mathbf{r}_i = g_{{align}}({\mathbf{z}}_i)$ for all $i$
\State Cluster $\{\mathbf{r}_i\}$ in $\mathcal{K}_i$ to initialize prototypes $\{\mu_m\}_{m=1}^{M}$

\For{epoch $e = 1$ to $E_2$}
    \For{each minibatch $\{(\mathbf{z}_i, y_i, \mathcal{K}_i)\}_{i=1}^B$}
        \State Compute aligned embeddings: $\mathbf{r}_i = g_{{align}}(\mathbf{z}_i)$
        \State Select nearest prototype to $\mathbf{r}_i$ from $\{\mu_m\}_{m=1}^M$
        \State Compute  $\mathcal{L}_{\text{cl}}$ and $\mathcal{L}_{\text{pcl}}$ 
        \State Compute total loss: $\mathcal{L} = (1{-}\lambda)\mathcal{L}_{\text{pcl}} + \lambda \mathcal{L}_{\text{cl}}$
        \State Update $g_{align}$  parameters via backpropagation
    \EndFor
\EndFor

\State Recompute $\mathbf{r}_i$ and update prototypes $\mu$ by clustering
\State \Return Optimized $g_{align}$ and prototypes $\{\mu_m\}_{m=1}^M$

\end{algorithmic}
\end{algorithm}

\subsection{Two-Stage Contrastive Alignment of DIF}

Algorithm~\ref{alg:dfams_train} delineates a two-stage contrastive training procedure designed to optimize DIF embeddings for robust alignment across disparate knowledge bases (KBs). In the first stage, DIF embeddings are extracted for each training instance by probing the pretrained LLM at the identified neurons, yielding a compact representation that encapsulates domain-specific traits. The initial phase, inter-KB alignment, leverages contrastive learning to encourage DIF embeddings from semantically similar content across different KBs to occupy proximate regions in the embedding space. This is achieved by repeatedly updating the alignment network $g_{align}$ to minimize the contrastive loss $\mathcal{L}_{cl}$, drawing together positive pairs and repelling negatives. The second stage, intra-KB alignment, further refines the DIF space by introducing prototype representations for subdomains or classes within each KB, initialized via unsupervised clustering. Here, a prototype-based contrastive loss $\mathcal{L}_{pcl}$ is incorporated alongside the global loss, jointly optimizing $g_{align}$ to produce discriminative representations that not only bridge domains but respect intra-domain structure. The training concludes with the computation of updated prototypes representative of key semantic clusters, returning both the optimized alignment network and the learned prototypes for later inference.

\subsection{Adaptive Prototype-Guided Routing}

Algorithm~\ref{alg:dfams_infer} proposes a dynamic routing strategy for inference, leveraging the previously learned prototypes to efficiently allocate retrieval resources based on the semantic relevance of incoming queries. Upon receiving a query, the system first computes its DIF embedding by probing the pretrained LLM at the selected neuron set, followed by transformation via the trained alignment encoder. The query embedding is then compared to all domain prototypes using a similarity metric, yielding a relevance score for each prototype. If the highest similarity score falls below a predefined threshold, the system abstains from retrieval, indicating insufficient semantic alignment. Otherwise, the algorithm selects the top-N most relevant prototypes and aggregates similarity scores by knowledge base, proportionally allocating retrieval slots according to their relative relevance. This prototype-guided mechanism enables adaptive, fine-grained routing decisions that efficiently direct retrieval attention to the most promising knowledge sources, facilitating both high precision and scalable inference in multi-domain settings.

\begin{algorithm}[H]
\caption{Adaptive Prototype-Guided Routing}
\label{alg:dfams_infer}
\begin{algorithmic}[1]
\Require Query $q$, trained encoder $g_{{align}}$, prototype set $\{\mu_m\}_{m=1}^M$, retrieval threshold $\tau$, top-$N$ selection size $N$, total slots $T$
\Ensure Routing weights $w_k$ for each knowledge base $k$
\State Compute DIF embedding: $\mathbf{z} \leftarrow \textsc{Probe}(\Theta(q), \mathcal{N})$
\State Compute aligned embedding: $\mathbf{q} \leftarrow g_{{align}}(\mathbf{z})$
\State Compute similarity scores: $s_i \leftarrow \text{sim}(\mathbf{q}, \mu_i)$ for all $i$
\If{$\max_i s_i < \tau$}
    \State \Return $\mathbf{0}$ \Comment{Abstain from retrieval}
\Else
    \State Identify top-$N$ prototypes: $\mathcal{I} \leftarrow \text{TopN}(s, N)$
    \For{each knowledge base $k$}
        \State Compute slot count for KB $k$:
        \[
        w_k \leftarrow \left\lfloor
        \frac{
        \sum_{i \in \mathcal{I},\,k_i=k} s_i
        }{
        \sum_{k'} \sum_{i \in \mathcal{I},\,k_i=k'} s_i
        } \cdot T
        \right\rfloor
        \]
    \EndFor
    \State \Return $[w_1, \dots, w_K]$
\EndIf
\end{algorithmic}
\end{algorithm}

\section{Retrieval Pipeline and Indexing Strategies}
\label{appendix:retrieval}
We adopt \textbf{FAISS}~\cite{douze2024faiss} for dense vector indexing across all retrieval settings. 
For the \textbf{Wikipedia (Wiki) knowledge base}, we follow the clustering strategy in RopMura~\cite{wu2025talk}, partitioning 1M English passages into 10 semantically coherent knowledge bases. Passages are first embedded using \texttt{Qwen-embedding-v2}~\cite{bai2023qwen} for clustering, and subsequently indexed with \texttt{all-MiniLM-L6-v2}~\cite{wang2020minilmv2}. 
For the \textbf{Medical (Med) knowledge base}~\cite{zhao2025medrag}, we replicate the knowledge base construction from RAGRoute~\cite{guerraoui2025efficient}, creating four distinct sources: PubMed, StatPearls, medical textbooks, and medical Wikipedia, each indexed using \texttt{all-ßMiniLM-L6-v2}. 
For the \textbf{Private Enterprise Policy (PEP) knowledge base}, which contains internal Chinese-language company documents spanning four sub-knowledge bases, we utilize the \texttt{GTE-base-zh} encoder~\cite{li2023towards} fr indexing. 
For each query, DFAMS retrieves the top-10 documents from the dynamically selected source(s).
\section{Dataset Construction and Sampling Strategies}
\label{appendix:datasets}
We conduct evaluations on three in-domain corpora and two out-of-domain (OOD) benchmarks.

\paragraph{Wiki Dataset Construction.}
The training set consists of 23,240 queries: among them, 2,100 queries explicitly require no retrieval, while 21,140 queries require retrieval of a single document from a single knowledge base.  We deliberately exclude queries involving cross-knowledge-base or cross-document multi-segment retrieval to reduce data construction complexity, which also aligns better with practical scenarios.  The test set contains 8,879 queries: 900 queries do not trigger retrieval, 969 queries require cross-knowledge-base multi-document collaboration, and 790 queries require same-knowledge-base multi-document collaboration.  This setup is designed to verify the robustness of our method in realistic multi-knowledge-base scenarios.

\paragraph{Med Dataset Construction.}
Following a similar processing pipeline as Wiki, we construct the training and test sets for the medical domain.  The training set includes only “no retrieval” or “single-knowledge-base single-segment” queries to reduce annotation costs.  The test set additionally incorporates “cross-knowledge-base multi-segment” and “same-knowledge-base multi-segment” queries to evaluate the system’s generalization ability in realistic multi-knowledge-base collaboration scenarios.  The test set contains 2975
samples, with 438 requiring cross-knowledge-base multi-document collaboration and 823 requiring same-knowledge-base multi-document collaboration, to assess robustness in real multi-knowledge-base environments.

\paragraph{PEP Dataset Construction.}
We evaluate FR capabilities on a private enterprise policy dataset: the training set contains 2,088 samples, all querying single company policy documents from one knowledge base.  The test set contains 344 samples, where queries are either categorized as “other” (requiring no retrieval) or require retrieval from a single knowledge base.  Since PEP doesn't have corresponding golden-label documents, recall statistics are not reported.

\paragraph{OOD Dataset Construction.}
Following the evaluation paradigm of RAGRoute \cite{guerraoui2025efficient}, we construct lightweight OOD test sets by extracting sub-questions from \textit{MMLU}~\cite{hendrycks2020measuring} and \textit{MIRAGE}~\cite{xiong2024benchmarking} that are most relevant to the topics covered by the existing four knowledge bases. For \textit{MMLU}, we retain 1,222 questions that are potentially related to the WIKI knowledge base; for \textit{MIRAGE}, we filter 1,546 open-domain QA samples with the highest entity co-occurrence with the four knowledge bases. Both subsets lack corresponding golden-label documents and are used solely to evaluate the model’s robustness and knowledge generalization under domain shift and non-retrieval conditions.

\section{Baseline Implementation Details}
\label{appendix:baselines}
We evaluate six representative methods under the DFAMS benchmark. 
The implementation details of these baseline methods are as follows:
\paragraph{No-RAG.} As a non-retrieval baseline, we directly apply the original LLM without any external knowledge. 
\paragraph{Merged-RAG.} There is no separation between individual knowledge bases — all content is integrated into a single, unified knowledge base and indexed together for retrieval.
\paragraph{Prompt.} A knowledge bases selection baseline where a powerful 70B teacher model is used to classify which knowledge bases should be retrieved. This is necessary because smaller models (e.g., 7B) exhibit poor performance on explicit knowledge bases routing tasks. The 70B model performs knowledge bases classification via prompt-based reasoning. Based on the selected corpora, relevant documents are retrieved and then passed to a 7B LLM for final answer generation.
\paragraph{CoT Prompt.} A knowledge bases selection where a 70B teacher model is used to perform corpus routing, but with chain-of-thought (CoT) prompting \cite{wei2022chain}.  Compared to Prompt, this variant enhances reasoning by explicitly incorporating intermediate steps during knowledge bases classification and downstream answer generation.
\paragraph{SFT.} A supervised fine-tuning baseline trained on the $D_{train}$ dataset using a cross-entropy loss. We fine-tuned two base models, Qwen2.5-7B and LLaMA3.1-8B, to predict the correct knowledge base for each query. The training specifically targets improving knowledge base selection accuracy.
\paragraph{RopMura.} A recent joint retrieval and routing method \cite{wu2025talk}. As our focus is on knowledge bases selection, we isolate and evaluate only the knowledge bases selection module. Multi-turn dialog components are disabled for fair comparison.
\paragraph{RAGRoute.} For each knowledge bases, we train an MLP-based router whose architecture and training settings exactly match those of our Align-MLP, ensuring a fair comparison \cite{guerraoui2025efficient}. All methods share the same encoder and retriever: we use \texttt{all-MiniLM-L6-v2} for Wiki and Med, and \texttt{gte-base-zh} for PEP.

\section{Metric Definitions and Evaluation Configuration}
\label{appendix:metrics}

We evaluate DFAMS using three complementary metrics:

\paragraph{Cls Acc.} This metric measures whether the method correctly identifies the relevant knowledge base(s). A prediction is considered correct if it matches the ground-truth KB in single-source cases, outputs \texttt{Others} when no retrieval is required, or fully covers all gold KBs in multi-source cases.

\paragraph{Recall.} This metric follows standard RAG evaluation and is computed over the Top-10 retrieved documents. It is calculated only for retrieval-triggering queries, and measures the proportion of gold documents that appear within the Top-10 retrieved results. Formally, it is defined as the number of retrieved gold documents divided by the total number of gold documents for a given query. Queries that do not require retrieval are excluded to better isolate and evaluate the retrieval component.

\paragraph{QA.} This metric evaluates the final response quality. For multiple-choice questions, we extract the predicted option(s) (e.g., A/B/C/D) from the model output and compare them against the ground-truth answer. For open-ended queries, responses are scored by an LLM-based judge~\cite{zheng2023judging} that assesses factual correctness and fluency. The score ranges from 0 to 10.

\section{Implementation Details}

To train the \textbf{Multi-Prototype Knowledge Alignment} module of DFAMS, we adopt a two-stage process. In the first stage, each private knowledge base independently extracts its own DIF representations from its local training data $\mathcal{D}_{\text{train}}$, where the activations of the selected neuron groups across designated layers are concatenated and pooled across tokens (e.g., using \texttt{AverageToken}) to form the final DIF representations. These DIFs capture the semantic characteristics of the local knowledge bases without exposing the underlying textual content, and are securely stored and transferred to the aligner for training. In the second stage, the DIF representations from all participating knowledge bases are aggregated and fed into the Align-MLP for joint optimization. The aligner is trained with an inter-KB supervised contrastive loss and an intra-KB multi-prototype contrastive loss (PCL), with prototypes initialized via KMeans clustering. This two-stage design effectively reduces redundant computations, improves overall training efficiency, and satisfies the data-locality and privacy-preservation requirements inherent to federated retrieval scenarios.

We train the Align$-$MLP aligner using the AdamW optimizer with a learning rate of $2\times10^{-4}$ and a cosine decay schedule. Unless otherwise specified, the batch size is set to 64, the temperature parameter to 0.07, and the number of contrastive learning (CL) epochs to four, during which only the CL objective is optimized. All models are trained for six epochs in total, with the final one to two epochs jointly optimizing both the CL and PCL. The DIF representations are extracted from the \texttt{mlp.up$\_$proj} components of the 26th and 27th transformer layers, where attribution-based analysis identifies neuron groups that are most relevant to the retrieval objective. Specifically, neurons are grouped in sets of 20, and the top 50 groups (corresponding to 1,000 neurons per layer) with the highest attribution scores are selected for DIF construction. The Align-MLP shares the same architecture as the probing network used in Probing-RAG, consisting of three fully connected layers with intermediate SiLU activations, layer normalization, and dropout regularization. The hidden dimension of the MLP is set to 512, and its input and output dimensions match those of the DIF representations to ensure alignment consistency.

We train the Align$-$MLP aligner using the AdamW optimizer with a learning rate of $2\times10^{-4}$ and a cosine decay schedule. Unless otherwise specified, the batch size is set to 64, the temperature parameter to 0.07, and the number of contrastive learning (CL) epochs to four, during which only the CL objective is optimized. All models are trained for six epochs in total, with the final one to two epochs jointly optimizing both the CL and PCL objectives. The DIF representations are extracted from the \texttt{mlp.up$\_$proj} components of the 26th and 27th transformer layers, where attribution-based analysis identifies neuron groups most relevant to the retrieval objective. Specifically, neurons are grouped in sets of 20, and the top 50 groups (corresponding to 1,000 neurons per layer) with the highest attribution scores are selected for DIF construction. The Align-MLP follows the same architecture as the probing network used in Probing-RAG, consisting of three fully connected layers with intermediate SiLU activations, layer normalization, and dropout regularization. The hidden dimension of the MLP is set to 512, and both its input and output dimensions are aligned with the 2,000-dimensional DIF feature space. Across all experiments, the loss weighting parameter is fixed to $\alpha=0.95$, meaning that the PCL term contributes 0.05 of the total loss. Other hyperparameter configurations are summarized in Table~\ref{tab:hyperparams}.

\begin{table}[t]
\centering
\small
\fontsize{7pt}{9pt}\selectfont
\begin{tabular}{lccccc}
\toprule
\textbf{Hyperparameter} & \textbf{Wiki} & \textbf{Med} & \textbf{PEP} & \textbf{MMLU} & \textbf{MIRAGE} \\
\midrule
Number of KBs & 10 & 4 & 4 & 10 & 4 \\
Learning Rate & 2e-4 & 1e-4 & 1e-4 & 2e-4 & 1e-4 \\
Epochs & 13 & 20 & 6 & 13 & 20 \\
CL Epochs & 12 & 22 & 4 & 12 & 22 \\
Batch Size & 64 & 64 & 64 & 64 & 64 \\
Threshold & 0.85 & 0.95 & 0.80 & 0.85 & 0.95 \\
\bottomrule
\end{tabular}
\caption{Hyperparameter summary across different datasets and experimental scenarios.}
\label{tab:hyperparams}
\end{table}

\section{Additional Results on Dynamic Information Flow}
\label{appendix:layer-selection}
\begin{figure*}[t]
  \includegraphics[width=0.48\linewidth]{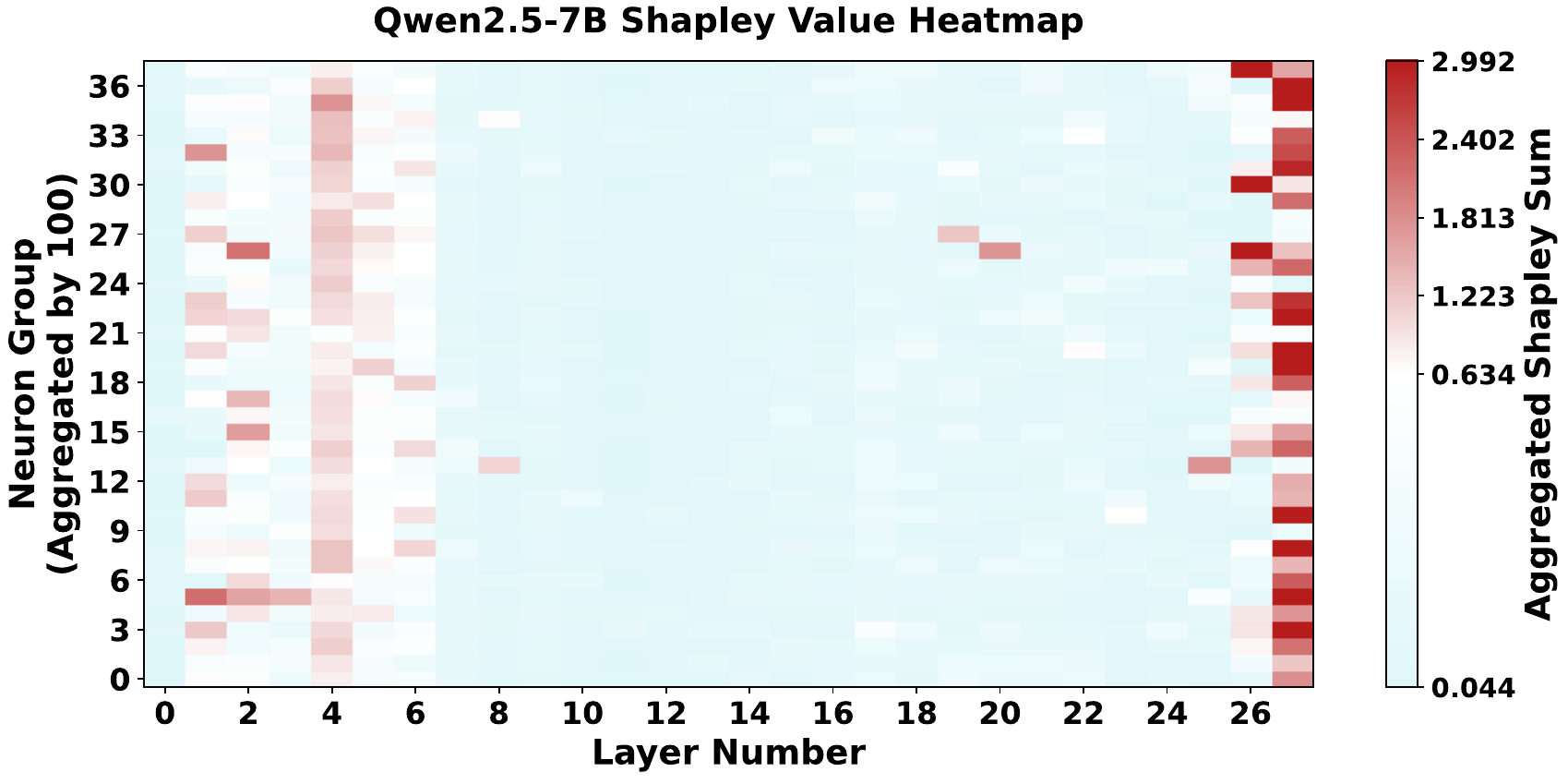} \hfill
  \includegraphics[width=0.48\linewidth]{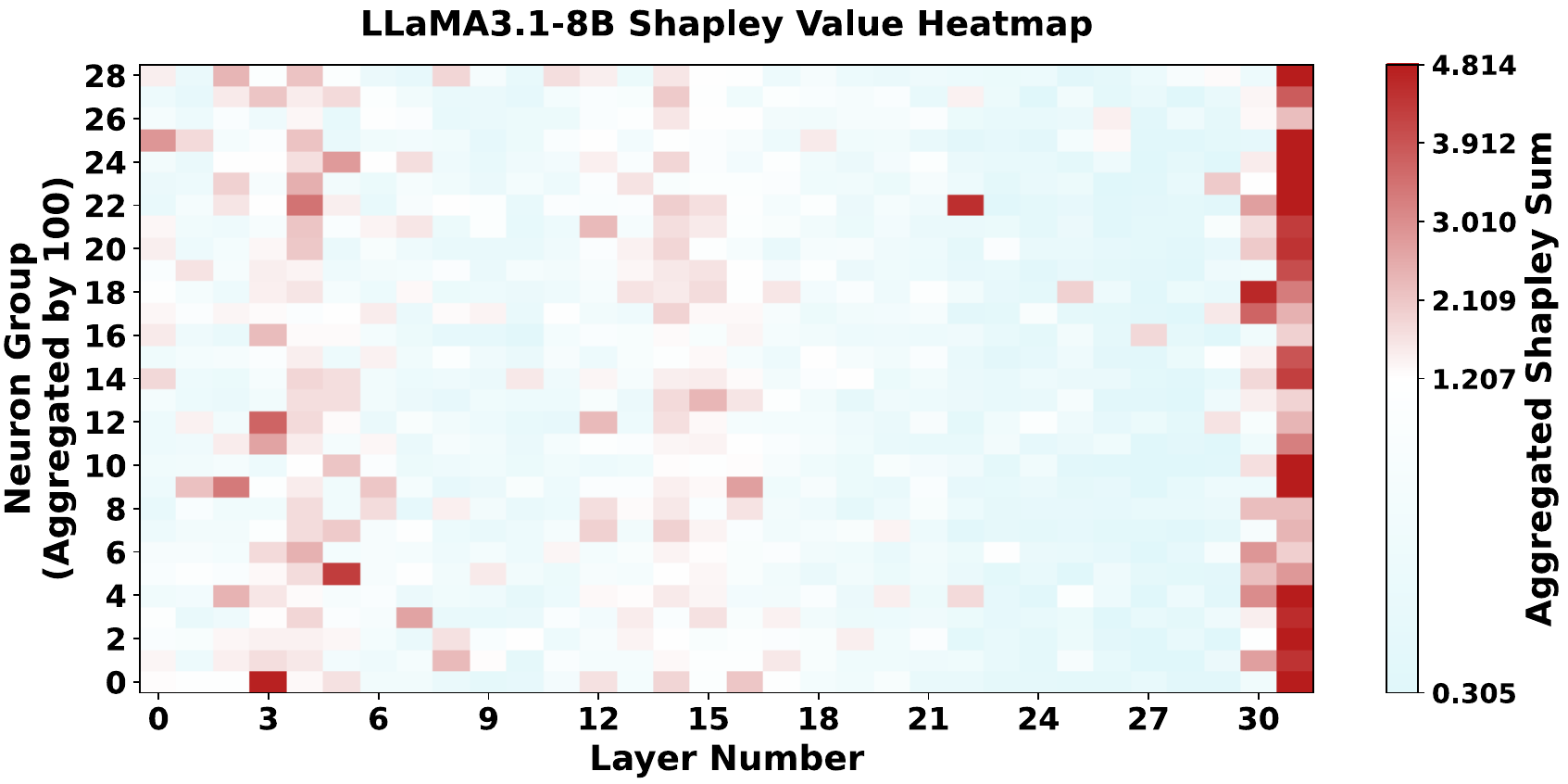} \hfill
\includegraphics[width=0.48\linewidth]{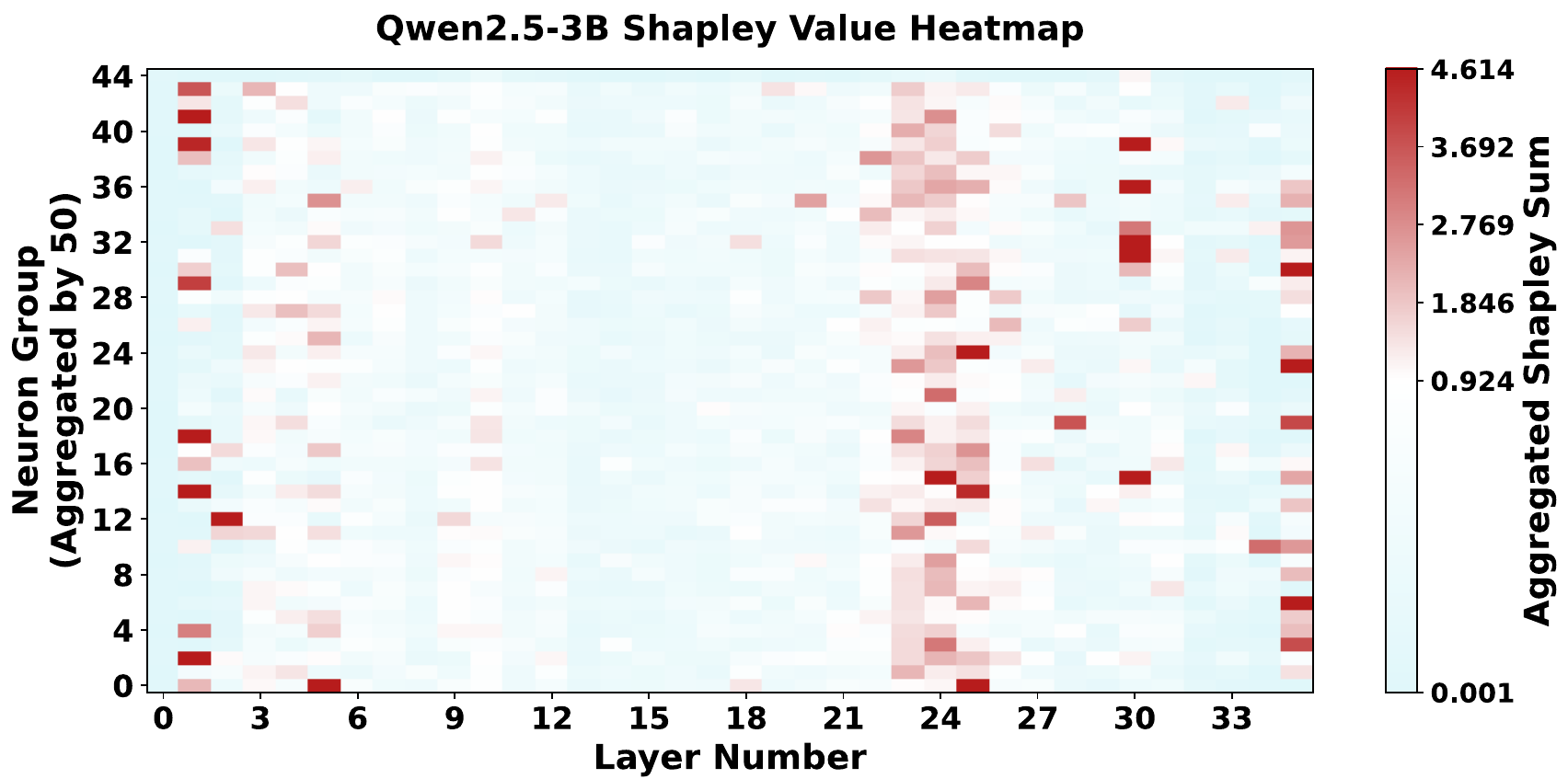} \hfill
  \includegraphics[width=0.48\linewidth]{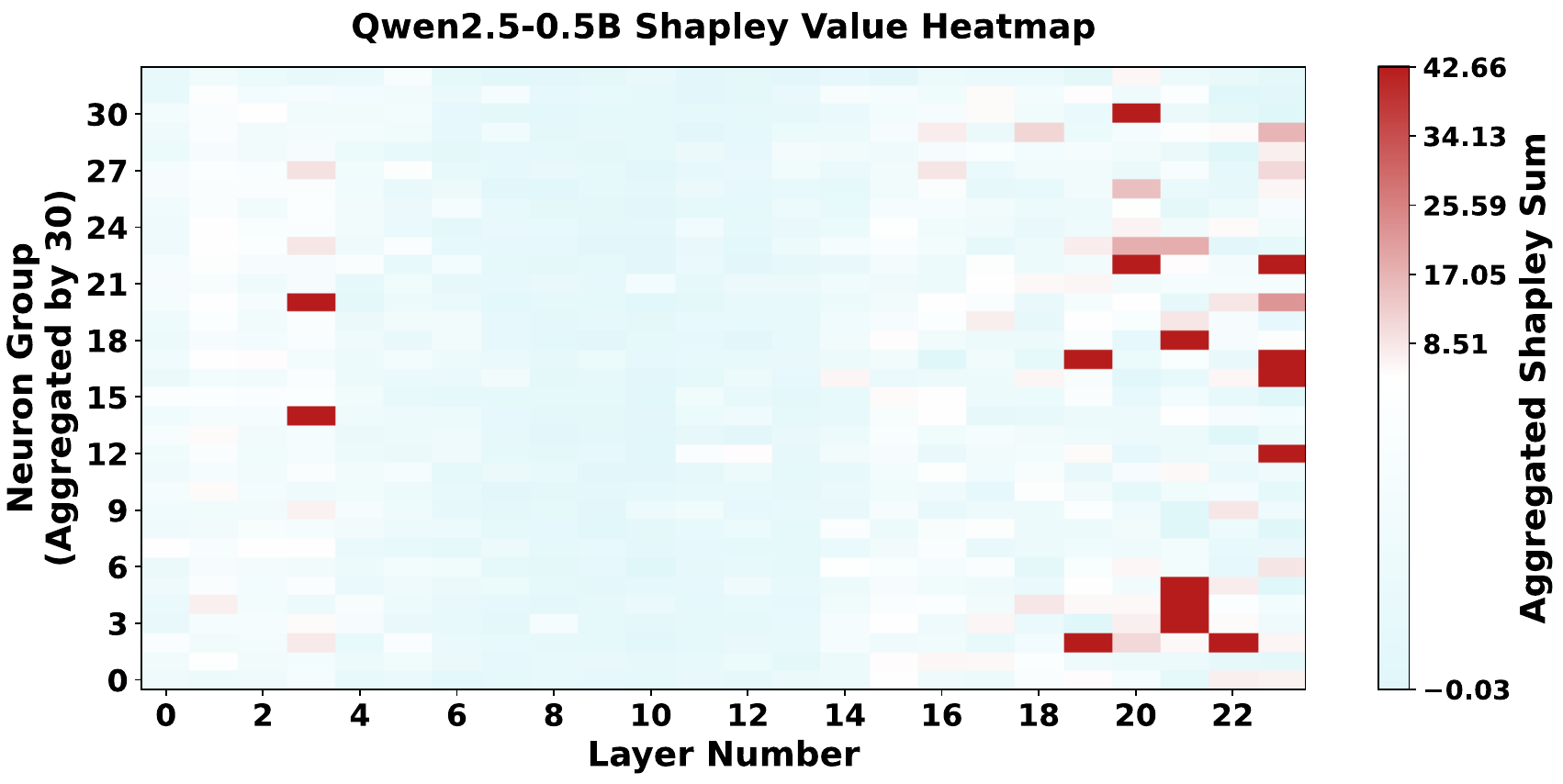}
  \caption {Heatmaps of Aggregated Shapley Values Across Layers and Neuron Groups for Qwen2.5-7B, Qwen2.5-3B, Qwen2.5-5B, and LLaMA3.1-8B Models}
  \label{fig:shapley_heatmap}
\end{figure*}

To verify our hypothesis regarding the presence of DIF, we analyzed the Shapley value heatmaps across four model sizes, as shown in Figure~\ref{fig:shapley_heatmap}. The results reveal a consistent trend: the Shapley values increase in the early-to-middle layers (e.g., around layers 1–6 in Qwen2.5-7B), decrease in the intermediate layers, and rise again in the deeper layers (e.g., layers 26–27). This pattern suggests that shallow layers primarily capture intent-related signals, while deeper layers activate and integrate domain-specific knowledge. This observation supports our decision to extract DIF representations from the latter layers.

We also conduct a layer-wise attribution analysis on the PEP dataset, as reported in Table~\ref{tab:layer_metrics}. Overall, the results show a generally positive correlation between Shapley magnitude and downstream performance—layers with higher Shapley values tend to achieve better classification accuracy. An exception is Layer 26, which attains slightly higher accuracy (72.70$\%$) despite a lower Shapley score (0.3765). In contrast, layers with small Shapley values, such as Layer 15 (0.0737), exhibit substantially lower accuracy (27.83$\%$). These observations confirm that Shapley-based analysis effectively highlights semantically informative layers contributing to domain-specific reasoning.
\begin{table}[ht]
\centering
\fontsize{7pt}{7pt}\selectfont
\renewcommand{\arraystretch}{1.2}
\setlength{\tabcolsep}{8pt}
\begin{tabular}{l|ccc|c}
\toprule
\rowcolor{gray!10}
\textbf{Layer} & \textbf{Act} & \textbf{Grad ($\times 10^{-6}$)} & \textbf{Shapley} & \textbf{Cls Acc} ($\uparrow$) \\
\midrule
26 & 7.53 & 0.671 &0.3765 & \textbf{72.70} \\
27 & -10.19 & 1.030 & \textbf{1.1895} & 72.51 \\
4  & 3.13 & 0.662 & 0.1357 & 56.61 \\
3  & 3.89 & 0.615 & 0.1339 & 46.17 \\
15 & 4.36 & 0.635 & 0.0737 & 27.83 \\
\bottomrule
\end{tabular}
\caption{Layer-wise attribution results on the PEP dataset.}
\label{tab:layer_metrics}
\end{table}

Finally, we compare three metrics—Shapley values, forward activations (Act)~\cite{xu2024parenting}, and gradients (Grad)~\cite{zhang2023adalora}—with gradient values scaled by $10^{-6}$ for readability. Among them, Shapley values most closely track model performance, whereas Act and Grad show weaker alignment. For instance, Layer 15 exhibits relatively high Act (4.36) and Grad (0.635) but performs poorly (27.83$\%$). These results underscore the superior reliability of Shapley-based attribution in identifying influential layers.

\section{Effect of Backbone Model Size.}
We evaluate DFAMS with \texttt{Qwen2.5} models of 0.5B, 3B, and 7B parameters. As shown in Table~\ref{tab:performance_on_wiki}, the 0.5B and 3B models achieve accuracy of 81.23\% and 83.56\%, and retrieval recall of 50.75\% and 52.12\%, respectively. Compared with the 7B model, the performance gap is relatively small, demonstrating that our framework can deliver strong performance even with smaller model sizes.

\begin{table}[ht]
\centering
\fontsize{9pt}{9pt}\selectfont
\renewcommand{\arraystretch}{1.2}
\setlength{\tabcolsep}{6pt}
\begin{tabular}{l|cc}
\toprule
\rowcolor{gray!10}
\textbf{Backbone} & \textbf{Cls Acc} ($\uparrow$) & \textbf{Recall} ($\uparrow$) \\
\midrule
Qwen2.5-0.5B & 81.23 & 50.75 \\
Qwen2.5-3B & 83.91 & 51.39 \\
Qwen2.5-7B & \textbf{85.03} & \textbf{53.83} \\
\bottomrule
\end{tabular}
\caption{Performance of DFAMS with different backbone models on the Wiki dataset.}
\label{tab:performance_on_wiki}
\end{table}

\section{Effect of Learning Rate}
Table~\ref{tab:lr} shows the classification accuracy of DFAMS under varying learning rates. The model achieves the best performance at a learning rate of \texttt{1e-4}, reaching 85.03\% accuracy. Both larger (\texttt{1e-3}: 81.51\%) and smaller (\texttt{1e-5}: 54.25\%) learning rates lead to performance drops.
\begin{table}[ht]
\centering
\fontsize{9pt}{9pt}\selectfont
\renewcommand{\arraystretch}{1.2}
\setlength{\tabcolsep}{6pt}
\begin{tabular}{l|ccccc}
\toprule
\rowcolor{gray!10}
Learning Rate & \texttt{1e-3} & \texttt{5e-4} & \texttt{1e-4} & \texttt{5e-5} & \texttt{1e-5} \\
\midrule
Cls Acc ($\uparrow$) & 81.51 & 83.56 & \textbf{85.03} & 83.22 & 54.25 \\
\bottomrule
\end{tabular}
\caption{Cls Acc of different learning rates using DFAMS.}
\label{tab:lr}
\end{table}


\section{Computational Resources and Software Environment}
Experiments were conducted on a server equipped with dual Intel Xeon E5-2680 v4 CPUs (56 cores, 112 threads), 8 NVIDIA RTX 3090 GPUs (24GB each), and 377 GB of main memory, running Ubuntu 18.04.6 LTS. Python 3.11.10 was used with PyTorch 2.4.0, and packages were managed via Conda 23.5.2. Model training took approximately 1 minute per epoch on average, depending on the experimental setting. Data preprocessing, including extraction of around 1,000 DIF samples (~4 MB), required about 10 minutes to obtain the DIF representations. During inference, generating QA accuracy results for 1,000 samples required approximately 5 hours on the same hardware configuration. All models and software packages, including HuggingFace Transformers 4.44.0, SpaCy 3.7.0, NLTK 3.8.1, and related dependencies, were used with default or explicitly stated parameter settings.

\section{The Use of Large Language Models}
In this work, Large Language Models (LLMs) were used for language polishing and coding assistance. Specifically, LLMs supported refining the clarity and grammar of the manuscript, improving stylistic quality, and suggesting code snippets or troubleshooting strategies. All content generated by LLMs was carefully reviewed and verified by the authors before inclusion.     The research design, critical analyses, and all final decisions were independently conducted by the authors. LLMs were not involved in generating new research ideas or conclusions.

\newpage
\section{Prompt}
\label{sec: prompts} 


In this section, we provide a detailed introduction to the prompts used in our framework:

\begin{tcolorbox}
[colback=lightgray!20,colframe=darkgray!80,title=Dataset Construction Prompt]
You are a knowledge expert tasked with creating a high-quality multiple-choice question based on the following text excerpt.

\textbf{Requirements:}
\begin{itemize}
  \item Question should be clear, concise.
  \item Provide four answer options A, B, C, and D.
  \item Only one correct answer; the other three must be plausible but incorrect.
  \item Answer must be directly supported by chunk.
  \item Output the result strictly in JSON format.
\end{itemize}

\textbf{Output Format:}
\begin{verbatim}
{
  "question": "Question content",
  "options": {
    "A": "Option A",
    "B": "Option B",
    "C": "Option C",
    "D": "Option D"
  },
  "answer": "Correct letter (A-D)"
}
\end{verbatim}
\textbf{Text excerpt:}

{text excerpt here (truncated to MAX\_TEXT\_LENGTH if needed)}
\end{tcolorbox}


\newpage

\begin{tcolorbox}
[colback=lightgray!20,colframe=darkgray!80,title=Multi-Chunk Dataset Construction Prompt]
You are an expert tasked with generating high-quality multiple-choice questions that integrates and synthesizes information across multiple chunks.

\textbf{Requirements:}
\begin{itemize}
  \item The question \textbf{must require synthesis of information from all {chunk\_num} text excerpts}. Avoid disjointed or unrelated pairings.
  \item The stem should naturally integrate ideas, characters, events, or facts from the various excerpts into a cohesive question.
  \item Do not generate a question that simply juxtaposes unrelated content from different texts — such questions are considered invalid.
  \item Ensure only one correct answer exists, and all distractors are plausible based on full context.
  \item If the question cannot reasonably be formed without being disjointed, return \texttt{false}.
  \item Return the result in \textbf{strict JSON format}.
\end{itemize}

\textbf{Output Format:}
\begin{verbatim}
{
  "question": "Question content",
  "options": {
    "A": "Option A",
    "B": "Option B",
    "C": "Option C",
    "D": "Option D"
  },
  "answer": "Correct letter (A-D)"
}
\end{verbatim}

Below are the {chunk\_num} related text excerpts. You must combine their information meaningfully in your question:

{text excerpt here (truncated to MAX\_TEXT\_LENGTH if needed)}
\end{tcolorbox}

\begin{tcolorbox}
[colback=lightgray!20,colframe=darkgray!80,title=DIF Probing Prompt]
\label{tab:adaptive_keyword_prompt}

You are a domain-specific large language model connected to the following knowledge bases:

\texttt{Knowledge Bases List:} \{database\_list\}

Given the query:

\texttt{"\{query\}"}

Please analyze and determine which knowledge base the query most likely belongs to. If it does not match any of the listed knowledge bases, respond with \texttt{others}.

\textbf{\textit{Response Format:}}

\texttt{Selected Knowledge Base:} [name of the knowledge base or \texttt{others}]
\end{tcolorbox}

\vspace{2em}

\begin{tcolorbox}
[colback=lightgray!20,colframe=darkgray!80,title=Prompt for Short Answer QA]
\label{tab:prompt_short_answer}
You are a professional QA assistant. Please answer the question based solely on the provided context. Follow the format below without omission:\\

\textbf{\textit{Prompt Format:}}

\texttt{<|im\_start|>system} \\
You are a professional QA assistant. Answer the question based on the given context. \\
\texttt{<|im\_end|>} \\

\texttt{<|im\_start|>user} \\
Context: \{context\} \\
Question: \{question\} \\
\texttt{<|im\_end|>} \\

\texttt{<|im\_start|>assistant} \\
Your answer here \\
\texttt{<|im\_end|>} 
\end{tcolorbox}

\newpage

\begin{tcolorbox}
[colback=lightgray!20,colframe=darkgray!80,title=Prompt for Multiple-Choice QA]
\label{tab:prompt_multiple_choice}
You are a professional multiple-choice QA assistant. Based on the provided context, answer the question by selecting the most appropriate option from A/B/C/D. Output only the option letter (A, B, C, or D) as the final answer; you may optionally add an explanation afterward.\\

\textbf{\textit{Prompt Format:}}

\texttt{<|im\_start|>system} \\
You are a professional multiple-choice QA assistant. Please answer the question based on the given context by selecting one option (A, B, C, or D). Output only the option letter as the final answer, optionally followed by an explanation. \\
\texttt{<|im\_end|>} \\

\texttt{<|im\_start|>user} \\
Context: \{context\} \\
Question: \{question\} \\
Options: \\
A. \{options.A\} \\
B. \{options.B\} \\
C. \{options.C\} \\
D. \{options.D\} \\

Please select the best option based on the above information. Output only the option letter, for example: "B" \\
\texttt{<|im\_end|>} \\

\texttt{<|im\_start|>assistant} \\
Your answer here \\
\texttt{<|im\_end|>} 
\end{tcolorbox}

\begin{tcolorbox}
[colback=lightgray!20,colframe=darkgray!80,title=Prompt for LLM Judgment of Open-Ended Answers]
\label{tab:prompt_evaluation}
You are a professional evaluator. Given the question, reference answer, and scoring criteria, please score the model-generated answer strictly from 0 to 10 (integer only). Return only the integer score without any extra text. \\

\textbf{\textit{Input:}}

\texttt{Question:} \{question\} \\
\texttt{Reference Answer:} \{standard\_answer\} \\
\texttt{Model Answer:} \{model\_answer\} \\

\textbf{\textit{Scoring Criteria:}}

1. Relevance: Does the answer directly address the question? (up to 4 points) \\
2. Accuracy: Is the content consistent with the reference answer? (up to 4 points) \\
3. Completeness: Does the answer cover key points in the reference? (up to 2 points) \\
4. Penalties: \\
\quad - Contains obvious errors: deduct 1–2 points \\
\quad - Completely unrelated or no answer: 0 points \\

\textbf{\textit{Examples:}} \\
- Perfect and complete: 10 points \\
- Mostly correct but missing some details: 8–9 points \\
- Partially correct: score proportionally (e.g., 3/5 key points = 6 points) \\
- Irrelevant but no errors: no deduction \\
- Completely wrong or no answer: 0 points \\

Please strictly follow the criteria and return only an integer score:
\end{tcolorbox}

\end{document}